\documentstyle{article}

\title{Better Language Models with Model Merging}

\author{Thorsten Brants\\
	Universit\"at des Saarlandes, Computational Linguistics\\
	P.O.Box 151150, D-66041 Saarbr\"ucken, Germany\\
	{\tt thorsten@coli.uni-sb.de}\\[2mm]
	{\em Conference on Empricial Methods in NLP}\\
	{\em May 17 -- 18, 1996, Philadelphia, PA.}}

\makeatletter
  \newlength\titlebox 
 \twocolumn 

\def\addcontentsline#1#2#3{}

\def\maketitle{\par
 \begingroup
   \def\thefootnote{\fnsymbol{footnote}}
   \def\@makefnmark{\hbox to 0pt{$^{\@thefnmark}$\hss}}
   \twocolumn[\@maketitle] \@thanks
 \endgroup
 \setcounter{footnote}{0}
 \let\maketitle\relax \let\@maketitle\relax
 \gdef\@thanks{}\gdef\@author{}\gdef\@title{}\let\thanks\relax}
\def\@maketitle{\vbox to \titlebox{\hsize\textwidth
 \linewidth\hsize \vskip 0.625in minus 0.125in \centering
 {\vspace{-0.5in}\LARGE\bf \@title \par} \vskip 0.2in plus 1fil minus 0.1in
 {\def\and{\unskip\enspace{\rm and}\enspace}%
  \def\And{\end{tabular}\hss \egroup \hskip 1in plus 2fil 
           \hbox to 0pt\bgroup\hss \begin{tabular}[t]{c}\Large\bf}%
  \def\AND{\end{tabular}\hss\egroup \hfil\hfil\egroup
	  \vskip 0.25in plus 1fil minus 0.125in
	   \hbox to \linewidth\bgroup\Large \hfil\hfil
 	     \hbox to 0pt\bgroup\hss \begin{tabular}[t]{c}\Large\bf}
  \hbox to \linewidth\bgroup\Large \hfil\hfil
    \hbox to 0pt\bgroup\hss \begin{tabular}[t]{c}\Large\bf\@author 
			    \end{tabular}\hss\egroup
    \hfil\hfil\egroup}
  \vskip 0.3in plus 2fil minus 0.1in
}}

\def\section{\@startsection {section}{1}{\z@}{-2.0ex plus
    -0.5ex minus -.2ex}{3pt plus 2pt minus 1pt}{\Large\bf\centering}}
\def\subsection{\@startsection{subsection}{2}{\z@}{-2.0ex plus
    -0.5ex minus -.2ex}{3pt plus 2pt minus 1pt}{\large\bf\raggedright}}
\def\subsubsection{\@startsection{subparagraph}{3}{\z@}{-6pt plus
   2pt minus 1pt}{-1em}{\normalsize\bf}}

\skip\footins 9pt plus 4pt minus 2pt
\def\footnoterule{\kern-3pt \hrule width 5pc \kern 2.6pt }
\labelwidth\leftmargini\advance\labelwidth-\labelsep \labelsep 5pt

\def\@listi{\leftmargin\leftmargini}
\def\@listii{\leftmargin\leftmarginii
   \labelwidth\leftmarginii\advance\labelwidth-\labelsep
   \topsep 2pt plus 1pt minus 0.5pt
   \parsep 1pt plus 0.5pt minus 0.5pt
   \itemsep \parsep}
\def\@listiii{\leftmargin\leftmarginiii
    \labelwidth\leftmarginiii\advance\labelwidth-\labelsep
    \topsep 1pt plus 0.5pt minus 0.5pt 
    \parsep \z@ \partopsep 0.5pt plus 0pt minus 0.5pt
    \itemsep \topsep}
\def\@listiv{\leftmargin\leftmarginiv
     \labelwidth\leftmarginiv\advance\labelwidth-\labelsep}
\def\@listv{\leftmargin\leftmarginv
     \labelwidth\leftmarginv\advance\labelwidth-\labelsep}
\def\@listvi{\leftmargin\leftmarginvi
     \labelwidth\leftmarginvi\advance\labelwidth-\labelsep}

\belowdisplayskip \abovedisplayskip
\def\@normalsize{\@setsize\normalsize{11pt}\xpt\@xpt}
\def\small{\@setsize\small{10pt}\ixpt\@ixpt}
\def\footnotesize{\@setsize\footnotesize{10pt}\ixpt\@ixpt}
\def\scriptsize{\@setsize\scriptsize{8pt}\viipt\@viipt}
\def\tiny{\@setsize\tiny{7pt}\vipt\@vipt}
\def\large{\@setsize\large{12pt}\xipt\@xipt}
\def\Large{\@setsize\Large{14pt}\xiipt\@xiipt}
\def\LARGE{\@setsize\LARGE{16pt}\xivpt\@xivpt}
\def\huge{\@setsize\huge{20pt}\xviipt\@xviipt}
\def\Huge{\@setsize\Huge{23pt}\xxpt\@xxpt}

\let\@internalcite\cite
\def\cite{\def\citename##1{##1, }\@internalcite}
\def\shortcite{\def\citename##1{}\@internalcite}
\def\newcite{\leavevmode\def\citename##1{{##1} (}\@internalciteb}

\def\@cite#1#2{({#1\if@tempswa , #2\fi})}

\catcode`@=11 \catcode`!=11
\expandafter\ifx\csname fiverm\endcsname\relax
  \let\fiverm\fivrm
\fi
\let\!latexendpicture=\endpicture 
\let\!latexframe=\frame
\let\!latexlinethickness=\linethickness
\let\!latexmultiput=\multiput
\let\!latexput=\put
\def\@picture(#1,#2)(#3,#4){%
  \@picht #2\unitlength
  \setbox\@picbox\hbox to #1\unitlength\bgroup 
  \let\endpicture=\!latexendpicture
  \let\frame=\!latexframe
  \let\linethickness=\!latexlinethickness
  \let\multiput=\!latexmultiput
  \let\put=\!latexput
  \hskip -#3\unitlength \lower #4\unitlength \hbox\bgroup}
\catcode`@=12 \catcode`!=12
\catcode`!=11 
\def\PiC{P\kern-.12em\lower.5ex\hbox{I}\kern-.075emC}
\def\PiCTeX{\PiC\kern-.11em\TeX}
\def\!ifnextchar#1#2#3{%
  \let\!testchar=#1%
  \def\!first{#2}%
  \def\!second{#3}%
  \futurelet\!nextchar\!testnext}
\def\!testnext{%
  \ifx \!nextchar \!spacetoken 
    \let\!next=\!skipspacetestagain
  \else
    \ifx \!nextchar \!testchar
      \let\!next=\!first
    \else 
      \let\!next=\!second 
    \fi 
  \fi
  \!next}
\def\\{\!skipspacetestagain} 
  \expandafter\def\\ {\futurelet\!nextchar\!testnext} 
\def\\{\let\!spacetoken= } \\  
\def\!tfor#1:=#2\do#3{%
  \edef\!fortemp{#2}%
  \ifx\!fortemp\!empty 
    \else
    \!tforloop#2\!nil\!nil\!!#1{#3}%
  \fi}
\def\!tforloop#1#2\!!#3#4{%
  \def#3{#1}%
  \ifx #3\!nnil
    \let\!nextwhile=\!fornoop
  \else
    #4\relax
    \let\!nextwhile=\!tforloop
  \fi 
  \!nextwhile#2\!!#3{#4}}
\def\!etfor#1:=#2\do#3{%
  \def\!!tfor{\!tfor#1:=}%
  \edef\!!!tfor{#2}%
  \expandafter\!!tfor\!!!tfor\do{#3}}
\def\!cfor#1:=#2\do#3{%
  \edef\!fortemp{#2}%
  \ifx\!fortemp\!empty 
  \else
    \!cforloop#2,\!nil,\!nil\!!#1{#3}%
  \fi}
\def\!cforloop#1,#2\!!#3#4{%
  \def#3{#1}%
  \ifx #3\!nnil
    \let\!nextwhile=\!fornoop 
  \else
    #4\relax
    \let\!nextwhile=\!cforloop
  \fi
  \!nextwhile#2\!!#3{#4}}
\def\!ecfor#1:=#2\do#3{%
  \def\!!cfor{\!cfor#1:=}%
  \edef\!!!cfor{#2}%
  \expandafter\!!cfor\!!!cfor\do{#3}}
\def\!empty{}
\def\!nnil{\!nil}
\def\!fornoop#1\!!#2#3{}
\def\!ifempty#1#2#3{%
  \edef\!emptyarg{#1}%
  \ifx\!emptyarg\!empty
    #2%
  \else
    #3%
  \fi}
\def\!getnext#1\from#2{%
  \expandafter\!gnext#2\!#1#2}%
\def\!gnext\\#1#2\!#3#4{%
  \def#3{#1}%
  \def#4{#2\\{#1}}%
  \ignorespaces}
\def\!getnextvalueof#1\from#2{%
  \expandafter\!gnextv#2\!#1#2}%
\def\!gnextv\\#1#2\!#3#4{%
  #3=#1%
  \def#4{#2\\{#1}}%
  \ignorespaces}
\def\!copylist#1\to#2{%
  \expandafter\!!copylist#1\!#2}
\def\!!copylist#1\!#2{%
  \def#2{#1}\ignorespaces}
\def\!wlet#1=#2{%
  \let#1=#2 
  \wlog{\string#1=\string#2}}
\def\!listaddon#1#2{%
  \expandafter\!!listaddon#2\!{#1}#2}
\def\!!listaddon#1\!#2#3{%
  \def#3{#1\\#2}}
\def\!rightappend#1\withCS#2\to#3{\expandafter\!!rightappend#3\!#2{#1}#3}
\def\!!rightappend#1\!#2#3#4{\def#4{#1#2{#3}}}
\def\!leftappend#1\withCS#2\to#3{\expandafter\!!leftappend#3\!#2{#1}#3}
\def\!!leftappend#1\!#2#3#4{\def#4{#2{#3}#1}}
\def\!lop#1\to#2{\expandafter\!!lop#1\!#1#2}
\def\!!lop\\#1#2\!#3#4{\def#4{#1}\def#3{#2}}
\def\!loop#1\repeat{\def\!body{#1}\!iterate}
\def\!iterate{\!body\let\!next=\!iterate\else\let\!next=\relax\fi\!next}
\def\!!loop#1\repeat{\def\!!body{#1}\!!iterate}
\def\!!iterate{\!!body\let\!!next=\!!iterate\else\let\!!next=\relax\fi\!!next}
\def\!removept#1#2{\edef#2{\expandafter\!!removePT\the#1}}
{\catcode`p=12 \catcode`t=12 \gdef\!!removePT#1pt{#1}}
\def\placevalueinpts of <#1> in #2 {%
  \!removept{#1}{#2}}
\def\!mlap#1{\hbox to 0pt{\hss#1\hss}}
\def\!vmlap#1{\vbox to 0pt{\vss#1\vss}}
\def\!not#1{%
  #1\relax
    \!switchfalse
  \else
    \!switchtrue
  \fi
  \if!switch
  \ignorespaces}
\let\!!!wlog=\wlog              
\def\wlog#1{}    
\newdimen\headingtoplotskip     
\newdimen\linethickness         
\newdimen\longticklength        
\newdimen\plotsymbolspacing     
\newdimen\shortticklength       
\newdimen\stackleading          
\newdimen\tickstovaluesleading  
\newdimen\totalarclength        
\newdimen\valuestolabelleading  
\newbox\!boxA                   
\newbox\!boxB                   
\newbox\!picbox                 
\newbox\!plotsymbol             
\newbox\!putobject              
\newbox\!shadesymbol            
\newcount\!countA               
\newcount\!countB               
\newcount\!countC               
\newcount\!countD               
\newcount\!countE               
\newcount\!countF               
\newcount\!countG               
\newcount\!fiftypt              
\newcount\!intervalno           
\newcount\!npoints              
\newcount\!nsegments            
\newcount\!ntemp                
\newcount\!parity               
\newcount\!scalefactor          
\newcount\!tfs                  
\newcount\!tickcase             
\newdimen\!Xleft                
\newdimen\!Xright               
\newdimen\!Xsave                
\newdimen\!Ybot                 
\newdimen\!Ysave                
\newdimen\!Ytop                 
\newdimen\!angle                
\newdimen\!arclength            
\newdimen\!areabloc             
\newdimen\!arealloc             
\newdimen\!arearloc             
\newdimen\!areatloc             
\newdimen\!bshrinkage           
\newdimen\!checkbot             
\newdimen\!checkleft            
\newdimen\!checkright           
\newdimen\!checktop             
\newdimen\!dimenA               
\newdimen\!dimenB               
\newdimen\!dimenC               
\newdimen\!dimenD               
\newdimen\!dimenE               
\newdimen\!dimenF               
\newdimen\!dimenG               
\newdimen\!dimenH               
\newdimen\!dimenI               
\newdimen\!distacross           
\newdimen\!downlength           
\newdimen\!dp                   
\newdimen\!dshade               
\newdimen\!dxpos                
\newdimen\!dxprime              
\newdimen\!dypos                
\newdimen\!dyprime              
\newdimen\!ht                   
\newdimen\!leaderlength         
\newdimen\!lshrinkage           
\newdimen\!midarclength         
\newdimen\!offset               
\newdimen\!plotheadingoffset    
\newdimen\!plotsymbolxshift     
\newdimen\!plotsymbolyshift     
\newdimen\!plotxorigin          
\newdimen\!plotyorigin          
\newdimen\!rootten              
\newdimen\!rshrinkage           
\newdimen\!shadesymbolxshift    
\newdimen\!shadesymbolyshift    
\newdimen\!tenAa                
\newdimen\!tenAc                
\newdimen\!tenAe                
\newdimen\!tshrinkage           
\newdimen\!uplength             
\newdimen\!wd                   
\newdimen\!wmax                 
\newdimen\!wmin                 
\newdimen\!xB                   
\newdimen\!xC                   
\newdimen\!xE                   
\newdimen\!xM                   
\newdimen\!xS                   
\newdimen\!xaxislength          
\newdimen\!xdiff                
\newdimen\!xleft                
\newdimen\!xloc                 
\newdimen\!xorigin              
\newdimen\!xpivot               
\newdimen\!xpos                 
\newdimen\!xprime               
\newdimen\!xright               
\newdimen\!xshade               
\newdimen\!xshift               
\newdimen\!xtemp                
\newdimen\!xunit                
\newdimen\!xxE                  
\newdimen\!xxM                  
\newdimen\!xxS                  
\newdimen\!xxloc                
\newdimen\!yB                   
\newdimen\!yC                   
\newdimen\!yE                   
\newdimen\!yM                   
\newdimen\!yS                   
\newdimen\!yaxislength          
\newdimen\!ybot                 
\newdimen\!ydiff                
\newdimen\!yloc                 
\newdimen\!yorigin              
\newdimen\!ypivot               
\newdimen\!ypos                 
\newdimen\!yprime               
\newdimen\!yshade               
\newdimen\!yshift               
\newdimen\!ytemp                
\newdimen\!ytop                 
\newdimen\!yunit                
\newdimen\!yyE                  
\newdimen\!yyM                  
\newdimen\!yyS                  
\newdimen\!yyloc                
\newdimen\!zpt                  
\newif\if!axisvisible           
\newif\if!gridlinestoo          
\newif\if!keepPO                
\newif\if!placeaxislabel        
\newif\if!switch                
\newif\if!xswitch               
\newtoks\!axisLaBeL             
\newtoks\!keywordtoks           
\newwrite\!replotfile           
\newhelp\!keywordhelp{The keyword mentioned in the error message in unknown. 
Replace NEW KEYWORD in the indicated response by the keyword that 
should have been specified.}    
\!wlet\!!origin=\!xM                   
\!wlet\!!unit=\!uplength               
\!wlet\!Lresiduallength=\!dimenG       
\!wlet\!Rresiduallength=\!dimenF       
\!wlet\!axisLength=\!distacross        
\!wlet\!axisend=\!ydiff                
\!wlet\!axisstart=\!xdiff              
\!wlet\!axisxlevel=\!arclength         
\!wlet\!axisylevel=\!downlength        
\!wlet\!beta=\!dimenE                  
\!wlet\!gamma=\!dimenF                 
\!wlet\!shadexorigin=\!plotxorigin     
\!wlet\!shadeyorigin=\!plotyorigin     
\!wlet\!ticklength=\!xS                
\!wlet\!ticklocation=\!xE              
\!wlet\!ticklocationincr=\!yE          
\!wlet\!tickwidth=\!yS                 
\!wlet\!totalleaderlength=\!dimenE     
\!wlet\!xone=\!xprime                  
\!wlet\!xtwo=\!dxprime                 
\!wlet\!ySsave=\!yM                    
\!wlet\!ybB=\!yB                       
\!wlet\!ybC=\!yC                       
\!wlet\!ybE=\!yE                       
\!wlet\!ybM=\!yM                       
\!wlet\!ybS=\!yS                       
\!wlet\!ybpos=\!yyloc                  
\!wlet\!yone=\!yprime                  
\!wlet\!ytB=\!xB                       
\!wlet\!ytC=\!xC                       
\!wlet\!ytE=\!downlength               
\!wlet\!ytM=\!arclength                
\!wlet\!ytS=\!distacross               
\!wlet\!ytpos=\!xxloc                  
\!wlet\!ytwo=\!dyprime                 
\!zpt=0pt                              
\!xunit=1pt
\!yunit=1pt
\!arearloc=\!xunit
\!areatloc=\!yunit
\!dshade=5pt
\!leaderlength=24in
\!tfs=256                              
\!wmax=5.3pt                           
\!wmin=2.7pt                           
\!xaxislength=\!xunit
\!xpivot=\!zpt
\!yaxislength=\!yunit 
\!ypivot=\!zpt
  \!dimenA=50pt \!fiftypt=\!dimenA     
\!rootten=3.162278pt                   
\!tenAa=8.690286pt                     
\!tenAc=2.773839pt                     
\!tenAe=2.543275pt                     
\def\!cosrotationangle{1}      
\def\!sinrotationangle{0}      
\def\!xpivotcoord{0}           
\def\!xref{0}                  
\def\!xshadesave{0}            
\def\!ypivotcoord{0}           
\def\!yref{0}                  
\def\!yshadesave{0}            
\def\!zero{0}                  
\let\wlog=\!!!wlog
\def\normalgraphs{%
  \longticklength=.4\baselineskip
  \shortticklength=.25\baselineskip
  \tickstovaluesleading=.25\baselineskip
  \valuestolabelleading=.8\baselineskip
  \linethickness=.4pt
  \stackleading=.17\baselineskip
  \headingtoplotskip=1.5\baselineskip
  \visibleaxes
  \ticksout
  \nogridlines
  \unloggedticks}
\def\setplotarea x from #1 to #2, y from #3 to #4 {%
  \!arealloc=\!M{#1}\!xunit \advance \!arealloc -\!xorigin
  \!areabloc=\!M{#3}\!yunit \advance \!areabloc -\!yorigin
  \!arearloc=\!M{#2}\!xunit \advance \!arearloc -\!xorigin
  \!areatloc=\!M{#4}\!yunit \advance \!areatloc -\!yorigin
  \!initinboundscheck
  \!xaxislength=\!arearloc  \advance\!xaxislength -\!arealloc
  \!yaxislength=\!areatloc  \advance\!yaxislength -\!areabloc
  \!plotheadingoffset=\!zpt
  \!dimenput {{\setbox0=\hbox{}\wd0=\!xaxislength\ht0=\!yaxislength\box0}}
     [bl] (\!arealloc,\!areabloc)}
\def\visibleaxes{%
  \def\!axisvisibility{\!axisvisibletrue}}

\def\!fixkeyword#1{%
  \errhelp=\!keywordhelp
  \errmessage{Unrecognized keyword `#1': \the\!keywordtoks{NEW KEYWORD}'}}
\!keywordtoks={enter `i\fixkeyword}
\def\fixkeyword#1{%
  \!nextkeyword#1 }
\def\axis {%
  \def\!nextkeyword##1 {%
    \expandafter\ifx\csname !axis##1\endcsname \relax
      \def\!next{\!fixkeyword{##1}}%
    \else
      \def\!next{\csname !axis##1\endcsname}%
    \fi
    \!next}%
  \!offset=\!zpt
  \!axisvisibility
  \!placeaxislabelfalse
  \!nextkeyword}
\def\!axisbottom{%
  \!axisylevel=\!areabloc
  \def\!tickxsign{0}%
  \def\!tickysign{-}%
  \def\!axissetup{\!axisxsetup}%
  \def\!axislabeltbrl{t}%
  \!nextkeyword}
\def\!axistop{%
  \!axisylevel=\!areatloc
  \def\!tickxsign{0}%
  \def\!tickysign{+}%
  \def\!axissetup{\!axisxsetup}%
  \def\!axislabeltbrl{b}%
  \!nextkeyword}
\def\!axisleft{%
  \!axisxlevel=\!arealloc
  \def\!tickxsign{-}%
  \def\!tickysign{0}%
  \def\!axissetup{\!axisysetup}%
  \def\!axislabeltbrl{r}%
  \!nextkeyword}
\def\!axisright{%
  \!axisxlevel=\!arearloc
  \def\!tickxsign{+}%
  \def\!tickysign{0}%
  \def\!axissetup{\!axisysetup}%
  \def\!axislabeltbrl{l}%
  \!nextkeyword}
\def\!axisshiftedto#1=#2 {%
  \if 0\!tickxsign
    \!axisylevel=\!M{#2}\!yunit
    \advance\!axisylevel -\!yorigin
  \else
    \!axisxlevel=\!M{#2}\!xunit
    \advance\!axisxlevel -\!xorigin
  \fi
  \!nextkeyword}
\def\!axisvisible{%
  \!axisvisibletrue  
  \!nextkeyword}
\def\!axisinvisible{%
  \!axisvisiblefalse
  \!nextkeyword}
\def\!axislabel#1 {%
  \!axisLaBeL={#1}%
  \!placeaxislabeltrue
  \!nextkeyword}
\expandafter\def\csname !axis/\endcsname{%
  \!axissetup 
  \if!placeaxislabel
    \!placeaxislabel
  \fi
  \if +\!tickysign 
    \!dimenA=\!axisylevel
    \advance\!dimenA \!offset 
    \advance\!dimenA -\!areatloc 
    \ifdim \!dimenA>\!plotheadingoffset
      \!plotheadingoffset=\!dimenA 
    \fi
  \fi}
\def\grid #1 #2 {%
  \!countA=#1\advance\!countA 1
  \axis bottom invisible ticks length <\!zpt> andacross quantity {\!countA} /
  \!countA=#2\advance\!countA 1
  \axis left   invisible ticks length <\!zpt> andacross quantity {\!countA} / }
\def\plotheading#1 {%
  \advance\!plotheadingoffset \headingtoplotskip
  \!dimenput {#1} [B] <.5\!xaxislength,\!plotheadingoffset>
    (\!arealloc,\!areatloc)}
\def\!axisxsetup{%
  \!axisxlevel=\!arealloc
  \!axisstart=\!arealloc
  \!axisend=\!arearloc
  \!axisLength=\!xaxislength
  \!!origin=\!xorigin
  \!!unit=\!xunit
  \!xswitchtrue
  \if!axisvisible 
    \!makeaxis
  \fi}
\def\!axisysetup{%
  \!axisylevel=\!areabloc
  \!axisstart=\!areabloc
  \!axisend=\!areatloc
  \!axisLength=\!yaxislength
  \!!origin=\!yorigin
  \!!unit=\!yunit
  \!xswitchfalse
  \if!axisvisible
    \!makeaxis
  \fi}
\def\!makeaxis{%
  \setbox\!boxA=\hbox{
    \beginpicture
      \!setdimenmode
      \setcoordinatesystem point at {\!zpt} {\!zpt}   
      \putrule from {\!zpt} {\!zpt} to
        {\!tickysign\!tickysign\!axisLength} 
        {\!tickxsign\!tickxsign\!axisLength}
    \endpicturesave <\!Xsave,\!Ysave>}%
    \wd\!boxA=\!zpt
    \!placetick\!axisstart}
\def\!placeaxislabel{%
  \advance\!offset \valuestolabelleading
  \if!xswitch
    \!dimenput {\the\!axisLaBeL} [\!axislabeltbrl]
      <.5\!axisLength,\!tickysign\!offset> (\!axisxlevel,\!axisylevel)
    \advance\!offset \!dp  
    \advance\!offset \!ht  
  \else
    \!dimenput {\the\!axisLaBeL} [\!axislabeltbrl]
      <\!tickxsign\!offset,.5\!axisLength> (\!axisxlevel,\!axisylevel)
  \fi
  \!axisLaBeL={}}
\def\arrow <#1> [#2,#3]{%
  \!ifnextchar<{\!arrow{#1}{#2}{#3}}{\!arrow{#1}{#2}{#3}<\!zpt,\!zpt> }}
\def\!arrow#1#2#3<#4,#5> from #6 #7 to #8 #9 {%
  \!xloc=\!M{#8}\!xunit   
  \!yloc=\!M{#9}\!yunit
  \!dxpos=\!xloc  \!dimenA=\!M{#6}\!xunit  \advance \!dxpos -\!dimenA
  \!dypos=\!yloc  \!dimenA=\!M{#7}\!yunit  \advance \!dypos -\!dimenA
  \let\!MAH=\!M
  \!setdimenmode
  \!xshift=#4\relax  \!yshift=#5\relax
  \!reverserotateonly\!xshift\!yshift
  \advance\!xshift\!xloc  \advance\!yshift\!yloc
  \!xS=-\!dxpos  \advance\!xS\!xshift
  \!yS=-\!dypos  \advance\!yS\!yshift
  \!start (\!xS,\!yS)
  \!ljoin (\!xshift,\!yshift)
  \!Pythag\!dxpos\!dypos\!arclength
  \!divide\!dxpos\!arclength\!dxpos  
  \!dxpos=32\!dxpos  \!removept\!dxpos\!!cos
  \!divide\!dypos\!arclength\!dypos  
  \!dypos=32\!dypos  \!removept\!dypos\!!sin
  \!halfhead{#1}{#2}{#3}
  \!halfhead{#1}{-#2}{-#3}
  \let\!M=\!MAH
  \ignorespaces}
  \def\!halfhead#1#2#3{%
    \!dimenC=-#1%
    \divide \!dimenC 2 
    \!dimenD=#2\!dimenC
    \!rotate(\!dimenC,\!dimenD)by(\!!cos,\!!sin)to(\!xM,\!yM)
    \!dimenC=-#1
    \!dimenD=#3\!dimenC
    \!dimenD=.5\!dimenD
    \!rotate(\!dimenC,\!dimenD)by(\!!cos,\!!sin)to(\!xE,\!yE)
    \!start (\!xshift,\!yshift)
    \advance\!xM\!xshift  \advance\!yM\!yshift
    \advance\!xE\!xshift  \advance\!yE\!yshift
    \!qjoin (\!xM,\!yM) (\!xE,\!yE) 
    \ignorespaces}
\def\betweenarrows #1#2 from #3 #4 to #5 #6 {%
  \!xloc=\!M{#3}\!xunit  \!xxloc=\!M{#5}\!xunit%
  \!yloc=\!M{#4}\!yunit  \!yyloc=\!M{#6}\!yunit%
  \!dxpos=\!xxloc  \advance\!dxpos by -\!xloc
  \!dypos=\!yyloc  \advance\!dypos by -\!yloc
  \advance\!xloc .5\!dxpos
  \advance\!yloc .5\!dypos
  \let\!MBA=\!M
  \!setdimenmode
  \ifdim\!dypos=\!zpt
    \ifdim\!dxpos<\!zpt \!dxpos=-\!dxpos \fi
    \put {\!lrarrows{\!dxpos}{#1}}#2{} at {\!xloc} {\!yloc}
  \else
    \ifdim\!dxpos=\!zpt
      \ifdim\!dypos<\!zpt \!dypos=-\!zpt \fi
      \put {\!udarrows{\!dypos}{#1}}#2{} at {\!xloc} {\!yloc}
    \fi
  \fi
  \let\!M=\!MBA
  \ignorespaces}
\def\!lrarrows#1#2{
  {\setbox\!boxA=\hbox{$\mkern-2mu\mathord-\mkern-2mu$}%
   \setbox\!boxB=\hbox{$\leftarrow$}\!dimenE=\ht\!boxB
   \setbox\!boxB=\hbox{}\ht\!boxB=2\!dimenE
   \hbox to #1{$\mathord\leftarrow\mkern-6mu
     \cleaders\copy\!boxA\hfil
     \mkern-6mu\mathord-$%
     \kern.4em $\vcenter{\box\!boxB}$$\vcenter{\hbox{#2}}$\kern.4em
     $\mathord-\mkern-6mu
     \cleaders\copy\!boxA\hfil
     \mkern-6mu\mathord\rightarrow$}}}
\def\!udarrows#1#2{
  {\setbox\!boxB=\hbox{#2}%
   \setbox\!boxA=\hbox to \wd\!boxB{\hss$\vert$\hss}%
   \!dimenE=\ht\!boxA \advance\!dimenE \dp\!boxA \divide\!dimenE 2
   \vbox to #1{\offinterlineskip
      \vskip .05556\!dimenE
      \hbox to \wd\!boxB{\hss$\mkern.4mu\uparrow$\hss}\vskip-\!dimenE
      \cleaders\copy\!boxA\vfil
      \vskip-\!dimenE\copy\!boxA
      \vskip\!dimenE\copy\!boxB\vskip.4em
      \copy\!boxA\vskip-\!dimenE
      \cleaders\copy\!boxA\vfil
      \vskip-\!dimenE \hbox to \wd\!boxB{\hss$\mkern.4mu\downarrow$\hss}
      \vskip .05556\!dimenE}}}
\def\putbar#1breadth <#2> from #3 #4 to #5 #6 {%
  \!xloc=\!M{#3}\!xunit  \!xxloc=\!M{#5}\!xunit%
  \!yloc=\!M{#4}\!yunit  \!yyloc=\!M{#6}\!yunit%
  \!dypos=\!yyloc  \advance\!dypos by -\!yloc
  \!dimenI=#2  
  \ifdim \!dimenI=\!zpt 
    \putrule#1from {#3} {#4} to {#5} {#6} 
  \else 
    \let\!MBar=\!M
    \!setdimenmode 
    \divide\!dimenI 2
    \ifdim \!dypos=\!zpt             
      \advance \!yloc -\!dimenI 
      \advance \!yyloc \!dimenI
    \else
      \advance \!xloc -\!dimenI 
      \advance \!xxloc \!dimenI
    \fi
    \putrectangle#1corners at {\!xloc} {\!yloc} and {\!xxloc} {\!yyloc}
    \let\!M=\!MBar 
  \fi
  \ignorespaces}
\def\setbars#1breadth <#2> baseline at #3 = #4 {%
  \edef\!barshift{#1}%
  \edef\!barbreadth{#2}%
  \edef\!barorientation{#3}%
  \edef\!barbaseline{#4}%
  \def\!bardobaselabel{\!bardoendlabel}%
  \def\!bardoendlabel{\!barfinish}%
  \let\!drawcurve=\!barcurve
  \!setbars}
\def\!setbars{%
  \futurelet\!nextchar\!!setbars}
\def\!!setbars{%
  \if b\!nextchar
    \def\!!!setbars{\!setbarsbget}%
  \else 
    \if e\!nextchar
      \def\!!!setbars{\!setbarseget}%
    \else
      \def\!!!setbars{\relax}%
    \fi
  \fi
  \!!!setbars}
\def\!setbarsbget baselabels (#1) {%
  \def\!barbaselabelorientation{#1}%
  \def\!bardobaselabel{\!!bardobaselabel}%
  \!setbars}
\def\!setbarseget endlabels (#1) {%
  \edef\!barendlabelorientation{#1}%
  \def\!bardoendlabel{\!!bardoendlabel}%
  \!setbars}
\def\!barcurve #1 #2 {%
  \if y\!barorientation
    \def\!basexarg{#1}%
    \def\!baseyarg{\!barbaseline}%
  \else
    \def\!basexarg{\!barbaseline}%
    \def\!baseyarg{#2}%
  \fi
  \expandafter\putbar\!barshift breadth <\!barbreadth> from {\!basexarg}
    {\!baseyarg} to {#1} {#2}
  \def\!endxarg{#1}%
  \def\!endyarg{#2}%
  \!bardobaselabel}
\def\!!bardobaselabel "#1" {%
  \put {#1}\!barbaselabelorientation{} at {\!basexarg} {\!baseyarg}
  \!bardoendlabel}
\def\!!bardoendlabel "#1" {%
  \put {#1}\!barendlabelorientation{} at {\!endxarg} {\!endyarg}
  \!barfinish}
\def\!barfinish{%
  \!ifnextchar/{\!finish}{\!barcurve}}
\def\putrectangle{%
  \!ifnextchar<{\!putrectangle}{\!putrectangle<\!zpt,\!zpt> }}
\def\!putrectangle<#1,#2> corners at #3 #4 and #5 #6 {%
  \!xone=\!M{#3}\!xunit  \!xtwo=\!M{#5}\!xunit%
  \!yone=\!M{#4}\!yunit  \!ytwo=\!M{#6}\!yunit%
  \ifdim \!xtwo<\!xone
    \!dimenI=\!xone  \!xone=\!xtwo  \!xtwo=\!dimenI
  \fi
  \ifdim \!ytwo<\!yone
    \!dimenI=\!yone  \!yone=\!ytwo  \!ytwo=\!dimenI
  \fi
  \!dimenI=#1\relax  \advance\!xone\!dimenI  \advance\!xtwo\!dimenI
  \!dimenI=#2\relax  \advance\!yone\!dimenI  \advance\!ytwo\!dimenI
  \let\!MRect=\!M
  \!setdimenmode
  \!shaderectangle
  \!dimenI=.5\linethickness
  \advance \!xone  -\!dimenI
  \advance \!xtwo   \!dimenI
  \putrule from {\!xone} {\!yone} to {\!xtwo} {\!yone} 
  \putrule from {\!xone} {\!ytwo} to {\!xtwo} {\!ytwo} 
  \advance \!xone   \!dimenI
  \advance \!xtwo  -\!dimenI%
  \advance \!yone  -\!dimenI
  \advance \!ytwo   \!dimenI
  \putrule from {\!xone} {\!yone} to {\!xone} {\!ytwo} 
  \putrule from {\!xtwo} {\!yone} to {\!xtwo} {\!ytwo} 
  \let\!M=\!MRect
  \ignorespaces}

\def\shaderectanglesoff{%
  \def\!shaderectangle{}%
  \ignorespaces}
\shaderectanglesoff
\def\!!shaderectangle{%
  \!dimenA=\!xtwo  \advance \!dimenA -\!xone
  \!dimenB=\!ytwo  \advance \!dimenB -\!yone
  \ifdim \!dimenA<\!dimenB
    \!startvshade (\!xone,\!yone,\!ytwo)
    \!lshade      (\!xtwo,\!yone,\!ytwo)
  \else
    \!starthshade (\!yone,\!xone,\!xtwo)
    \!lshade      (\!ytwo,\!xone,\!xtwo)
  \fi
  \ignorespaces}
\def\frame{%
  \!ifnextchar<{\!frame}{\!frame<\!zpt> }}
\long\def\!frame<#1> #2{%
  \beginpicture
    \setcoordinatesystem units <1pt,1pt> point at 0 0 
    \put {#2} [Bl] at 0 0 
    \!dimenA=#1\relax
    \!dimenB=\!wd \advance \!dimenB \!dimenA
    \!dimenC=\!ht \advance \!dimenC \!dimenA
    \!dimenD=\!dp \advance \!dimenD \!dimenA
    \let\!MFr=\!M
    \!setdimenmode
    \putrectangle corners at {-\!dimenA} {-\!dimenD} and {\!dimenB} {\!dimenC}
    \!setcoordmode
    \let\!M=\!MFr
  \endpicture
  \ignorespaces}
\def\rectangle <#1> <#2> {%
  \setbox0=\hbox{}\wd0=#1\ht0=#2\frame {\box0}}
\def\plot{%
  \!ifnextchar"{\!plotfromfile}{\!drawcurve}}
\def\!plotfromfile"#1"{%
  \expandafter\!drawcurve \input #1 /}
\def\setquadratic{%
  \let\!drawcurve=\!qcurve
  \let\!!Shade=\!!qShade
  \let\!!!Shade=\!!!qShade}
\def\setlinear{%
  \let\!drawcurve=\!lcurve
  \let\!!Shade=\!!lShade
  \let\!!!Shade=\!!!lShade}
\def\sethistograms{%
  \let\!drawcurve=\!hcurve}
\def\!qcurve #1 #2 {%
  \!start (#1,#2)
  \!Qjoin}
\def\!Qjoin#1 #2 #3 #4 {%
  \!qjoin (#1,#2) (#3,#4)             
  \!ifnextchar/{\!finish}{\!Qjoin}}
\def\!lcurve #1 #2 {%
  \!start (#1,#2)
  \!Ljoin}
\def\!Ljoin#1 #2 {%
  \!ljoin (#1,#2)                    
  \!ifnextchar/{\!finish}{\!Ljoin}}
\def\!finish/{\ignorespaces}
\def\!hcurve #1 #2 {%
  \edef\!hxS{#1}%
  \edef\!hyS{#2}%
  \!hjoin}
\def\!hjoin#1 #2 {%
  \putrectangle corners at {\!hxS} {\!hyS} and {#1} {#2}
  \edef\!hxS{#1}%
  \!ifnextchar/{\!finish}{\!hjoin}}
\def\vshade #1 #2 #3 {%
  \!startvshade (#1,#2,#3)
  \!Shadewhat}
\def\hshade #1 #2 #3 {%
  \!starthshade (#1,#2,#3)
  \!Shadewhat}
\def\!Shadewhat{%
  \futurelet\!nextchar\!Shade}
\def\!Shade{%
  \if <\!nextchar
    \def\!nextShade{\!!Shade}%
  \else
    \if /\!nextchar
      \def\!nextShade{\!finish}%
    \else
      \def\!nextShade{\!!!Shade}%
    \fi
  \fi
  \!nextShade}
\def\!!lShade<#1> #2 #3 #4 {%
  \!lshade <#1> (#2,#3,#4)                 
  \!Shadewhat}
\def\!!!lShade#1 #2 #3 {%
  \!lshade (#1,#2,#3)
  \!Shadewhat} 
\def\!!qShade<#1> #2 #3 #4 #5 #6 #7 {%
  \!qshade <#1> (#2,#3,#4) (#5,#6,#7)      
  \!Shadewhat}
\def\!!!qShade#1 #2 #3 #4 #5 #6 {%
  \!qshade (#1,#2,#3) (#4,#5,#6)
  \!Shadewhat} 
\setlinear
\def\setdashpattern <#1>{%
  \def\!Flist{}\def\!Blist{}\def\!UDlist{}%
  \!countA=0
  \!ecfor\!item:=#1\do{%
    \!dimenA=\!item\relax
    \expandafter\!rightappend\the\!dimenA\withCS{\\}\to\!UDlist%
    \advance\!countA  1
    \ifodd\!countA
      \expandafter\!rightappend\the\!dimenA\withCS{\!Rule}\to\!Flist%
      \expandafter\!leftappend\the\!dimenA\withCS{\!Rule}\to\!Blist%
    \else 
      \expandafter\!rightappend\the\!dimenA\withCS{\!Skip}\to\!Flist%
      \expandafter\!leftappend\the\!dimenA\withCS{\!Skip}\to\!Blist%
    \fi}%
  \!leaderlength=\!zpt
  \def\!Rule##1{\advance\!leaderlength  ##1}%
  \def\!Skip##1{\advance\!leaderlength  ##1}%
  \!Flist%
  \ifdim\!leaderlength>\!zpt 
  \else
    \def\!Flist{\!Skip{24in}}\def\!Blist{\!Skip{24in}}\ignorespaces
    \def\!UDlist{\\{\!zpt}\\{24in}}\ignorespaces
    \!leaderlength=24in
  \fi
  \!dashingon}   
\def\!dashingon{%
  \def\!advancedashing{\!!advancedashing}%
  \def\!drawlinearsegment{\!lineardashed}%
  \def\!puthline{\!putdashedhline}%
  \def\!putvline{\!putdashedvline}%
  \ignorespaces}%
\def\!dashingoff{%
  \def\!advancedashing{\relax}%
  \def\!drawlinearsegment{\!linearsolid}%
  \def\!puthline{\!putsolidhline}%
  \def\!putvline{\!putsolidvline}%
  \ignorespaces}
\def\setdots{%
  \!ifnextchar<{\!setdots}{\!setdots<5pt>}}
\def\!setdots<#1>{%
  \!dimenB=#1\advance\!dimenB -\plotsymbolspacing
  \ifdim\!dimenB<\!zpt
    \!dimenB=\!zpt
  \fi
\setdashpattern <\plotsymbolspacing,\!dimenB>}
\def\setdotsnear <#1> for <#2>{%
  \!dimenB=#2\relax  \advance\!dimenB -.05pt  
  \!dimenC=#1\relax  \!countA=\!dimenC 
  \!dimenD=\!dimenB  \advance\!dimenD .5\!dimenC  \!countB=\!dimenD
  \divide \!countB  \!countA
  \ifnum 1>\!countB 
    \!countB=1
  \fi
  \divide\!dimenB  \!countB
  \setdots <\!dimenB>}
\def\setdashes{%
  \!ifnextchar<{\!setdashes}{\!setdashes<5pt>}}
\def\!setdashes<#1>{\setdashpattern <#1,#1>}
\def\setdashesnear <#1> for <#2>{%
  \!dimenB=#2\relax  
  \!dimenC=#1\relax  \!countA=\!dimenC 
  \!dimenD=\!dimenB  \advance\!dimenD .5\!dimenC  \!countB=\!dimenD
  \divide \!countB  \!countA
  \ifodd \!countB 
  \else 
    \advance \!countB  1
  \fi
  \divide\!dimenB  \!countB
  \setdashes <\!dimenB>}
\def\setsolid{%
  \def\!Flist{\!Rule{24in}}\def\!Blist{\!Rule{24in}}%
  \def\!UDlist{\\{24in}\\{\!zpt}}%
  \!dashingoff}  
\setsolid

\def\!divide#1#2#3{%
  \!dimenB=#1
  \!dimenC=#2
  \!dimenD=\!dimenB
  \divide \!dimenD \!dimenC
  \!dimenA=\!dimenD
  \multiply\!dimenD \!dimenC
  \advance\!dimenB -\!dimenD
  \!dimenD=\!dimenC
    \ifdim\!dimenD<\!zpt \!dimenD=-\!dimenD 
  \fi
  \ifdim\!dimenD<64pt
    \!divstep[\!tfs]\!divstep[\!tfs]%
  \else 
    \!!divide
  \fi
  #3=\!dimenA\ignorespaces}
\def\!!divide{%
  \ifdim\!dimenD<256pt
    \!divstep[64]\!divstep[32]\!divstep[32]%
  \else 
    \!divstep[8]\!divstep[8]\!divstep[8]\!divstep[8]\!divstep[8]%
    \!dimenA=2\!dimenA
  \fi}
\def\!divstep[#1]{
  \!dimenB=#1\!dimenB
  \!dimenD=\!dimenB
    \divide \!dimenD by \!dimenC
  \!dimenA=#1\!dimenA
    \advance\!dimenA by \!dimenD%
  \multiply\!dimenD by \!dimenC
    \advance\!dimenB by -\!dimenD}
\def\Divide <#1> by <#2> forming <#3> {%
  \!divide{#1}{#2}{#3}}

\def\ellipticalarc axes ratio #1:#2 #3 degrees from #4 #5 center at #6 #7 {%
  \!angle=#3pt\relax
  \ifdim\!angle>\!zpt 
    \def\!sign{}
  \else 
    \def\!sign{-}\!angle=-\!angle
  \fi
  \!xxloc=\!M{#6}\!xunit
  \!yyloc=\!M{#7}\!yunit     
  \!xxS=\!M{#4}\!xunit
  \!yyS=\!M{#5}\!yunit
  \advance\!xxS -\!xxloc
  \advance\!yyS -\!yyloc
  \!divide\!xxS{#1pt}\!xxS 
  \!divide\!yyS{#2pt}\!yyS 
  \let\!MC=\!M
  \!setdimenmode
  \!xS=#1\!xxS  \advance\!xS\!xxloc
  \!yS=#2\!yyS  \advance\!yS\!yyloc
  \!start (\!xS,\!yS)%
  \!loop\ifdim\!angle>14.9999pt
    \!rotate(\!xxS,\!yyS)by(\!cos,\!sign\!sin)to(\!xxM,\!yyM) 
    \!rotate(\!xxM,\!yyM)by(\!cos,\!sign\!sin)to(\!xxE,\!yyE)
    \!xM=#1\!xxM  \advance\!xM\!xxloc  \!yM=#2\!yyM  \advance\!yM\!yyloc
    \!xE=#1\!xxE  \advance\!xE\!xxloc  \!yE=#2\!yyE  \advance\!yE\!yyloc
    \!qjoin (\!xM,\!yM) (\!xE,\!yE)
    \!xxS=\!xxE  \!yyS=\!yyE 
    \advance \!angle -15pt
  \repeat
  \ifdim\!angle>\!zpt
    \!angle=100.53096\!angle
    \divide \!angle 360 
    \!sinandcos\!angle\!!sin\!!cos
    \!rotate(\!xxS,\!yyS)by(\!!cos,\!sign\!!sin)to(\!xxM,\!yyM) 
    \!rotate(\!xxM,\!yyM)by(\!!cos,\!sign\!!sin)to(\!xxE,\!yyE)
    \!xM=#1\!xxM  \advance\!xM\!xxloc  \!yM=#2\!yyM  \advance\!yM\!yyloc
    \!xE=#1\!xxE  \advance\!xE\!xxloc  \!yE=#2\!yyE  \advance\!yE\!yyloc
    \!qjoin (\!xM,\!yM) (\!xE,\!yE)
  \fi
  \let\!M=\!MC
  \ignorespaces}
\def\!rotate(#1,#2)by(#3,#4)to(#5,#6){%
  \!dimenA=#3#1\advance \!dimenA -#4#2
  \!dimenB=#3#2\advance \!dimenB  #4#1
  \divide \!dimenA 32  \divide \!dimenB 32 
  #5=\!dimenA  #6=\!dimenB
  \ignorespaces}
\def\!sin{4.17684}
\def\!cos{31.72624}
\def\!sinandcos#1#2#3{%
 \!dimenD=#1
 \!dimenA=\!dimenD
 \!dimenB=32pt
 \!removept\!dimenD\!value
 \!dimenC=\!dimenD
 \!dimenC=\!value\!dimenC \divide\!dimenC by 64 
 \advance\!dimenB by -\!dimenC
 \!dimenC=\!value\!dimenC \divide\!dimenC by 96 
 \advance\!dimenA by -\!dimenC
 \!dimenC=\!value\!dimenC \divide\!dimenC by 128 
 \advance\!dimenB by \!dimenC%
 \!removept\!dimenA#2
 \!removept\!dimenB#3
 \ignorespaces}
\def\putrule#1from #2 #3 to #4 #5 {%
  \!xloc=\!M{#2}\!xunit  \!xxloc=\!M{#4}\!xunit%
  \!yloc=\!M{#3}\!yunit  \!yyloc=\!M{#5}\!yunit%
  \!dxpos=\!xxloc  \advance\!dxpos by -\!xloc
  \!dypos=\!yyloc  \advance\!dypos by -\!yloc
  \ifdim\!dypos=\!zpt
    \def\!!Line{\!puthline{#1}}\ignorespaces
  \else
    \ifdim\!dxpos=\!zpt
      \def\!!Line{\!putvline{#1}}\ignorespaces
    \else 
       \def\!!Line{}
    \fi
  \fi
  \let\!ML=\!M
  \!setdimenmode
  \!!Line%
  \let\!M=\!ML
  \ignorespaces}
\def\!putsolidhline#1{%
  \ifdim\!dxpos>\!zpt 
    \put{\!hline\!dxpos}#1[l] at {\!xloc} {\!yloc}
  \else 
    \put{\!hline{-\!dxpos}}#1[l] at {\!xxloc} {\!yyloc}
  \fi
  \ignorespaces}
\def\!putsolidvline#1{%
  \ifdim\!dypos>\!zpt 
    \put{\!vline\!dypos}#1[b] at {\!xloc} {\!yloc}
  \else 
    \put{\!vline{-\!dypos}}#1[b] at {\!xxloc} {\!yyloc}
  \fi
  \ignorespaces}
\def\!hline#1{\hbox to #1{\leaders \hrule height\linethickness\hfill}}
\def\!vline#1{\vbox to #1{\leaders \vrule width\linethickness\vfill}}
\def\!putdashedhline#1{%
  \ifdim\!dxpos>\!zpt 
    \!DLsetup\!Flist\!dxpos
    \put{\hbox to \!totalleaderlength{\!hleaders}\!hpartialpattern\!Rtrunc}
      #1[l] at {\!xloc} {\!yloc} 
  \else 
    \!DLsetup\!Blist{-\!dxpos}
    \put{\!hpartialpattern\!Ltrunc\hbox to \!totalleaderlength{\!hleaders}}
      #1[r] at {\!xloc} {\!yloc} 
  \fi
  \ignorespaces}
\def\!putdashedvline#1{%
  \!dypos=-\!dypos
  \ifdim\!dypos>\!zpt 
    \!DLsetup\!Flist\!dypos 
    \put{\vbox{\vbox to \!totalleaderlength{\!vleaders}
      \!vpartialpattern\!Rtrunc}}#1[t] at {\!xloc} {\!yloc} 
  \else 
    \!DLsetup\!Blist{-\!dypos}
    \put{\vbox{\!vpartialpattern\!Ltrunc
      \vbox to \!totalleaderlength{\!vleaders}}}#1[b] at {\!xloc} {\!yloc} 
  \fi
  \ignorespaces}
\def\!DLsetup#1#2{
  \let\!RSlist=#1
  \!countB=#2
  \!countA=\!leaderlength
  \divide\!countB by \!countA
  \!totalleaderlength=\!countB\!leaderlength
  \!Rresiduallength=#2%
  \advance \!Rresiduallength by -\!totalleaderlength
  \!Lresiduallength=\!leaderlength
  \advance \!Lresiduallength by -\!Rresiduallength
  \ignorespaces}
\def\!hleaders{%
  \def\!Rule##1{\vrule height\linethickness width##1}%
  \def\!Skip##1{\hskip##1}%
  \leaders\hbox{\!RSlist}\hfill}
\def\!hpartialpattern#1{%
  \!dimenA=\!zpt \!dimenB=\!zpt 
  \def\!Rule##1{#1{##1}\vrule height\linethickness width\!dimenD}%
  \def\!Skip##1{#1{##1}\hskip\!dimenD}%
  \!RSlist}
\def\!vleaders{%
  \def\!Rule##1{\hrule width\linethickness height##1}%
  \def\!Skip##1{\vskip##1}%
  \leaders\vbox{\!RSlist}\vfill}
\def\!vpartialpattern#1{%
  \!dimenA=\!zpt \!dimenB=\!zpt 
  \def\!Rule##1{#1{##1}\hrule width\linethickness height\!dimenD}%
  \def\!Skip##1{#1{##1}\vskip\!dimenD}%
  \!RSlist}
\def\!Rtrunc#1{\!trunc{#1}>\!Rresiduallength}
\def\!Ltrunc#1{\!trunc{#1}<\!Lresiduallength}
\def\!trunc#1#2#3{%
  \!dimenA=\!dimenB         
  \advance\!dimenB by #1%
  \!dimenD=\!dimenB  \ifdim\!dimenD#2#3\!dimenD=#3\fi
  \!dimenC=\!dimenA  \ifdim\!dimenC#2#3\!dimenC=#3\fi
  \advance \!dimenD by -\!dimenC}
\def\!start (#1,#2){%
  \!plotxorigin=\!xorigin  \advance \!plotxorigin by \!plotsymbolxshift
  \!plotyorigin=\!yorigin  \advance \!plotyorigin by \!plotsymbolyshift
  \!xS=\!M{#1}\!xunit \!yS=\!M{#2}\!yunit
  \!rotateaboutpivot\!xS\!yS
  \!copylist\!UDlist\to\!!UDlist
  \!getnextvalueof\!downlength\from\!!UDlist
  \!distacross=\!zpt
  \!intervalno=0 
  \global\totalarclength=\!zpt
  \ignorespaces}
\def\!ljoin (#1,#2){%
  \advance\!intervalno by 1
  \!xE=\!M{#1}\!xunit \!yE=\!M{#2}\!yunit
  \!rotateaboutpivot\!xE\!yE
  \!xdiff=\!xE \advance \!xdiff by -\!xS
  \!ydiff=\!yE \advance \!ydiff by -\!yS
  \!Pythag\!xdiff\!ydiff\!arclength
  \global\advance \totalarclength by \!arclength%
  \!drawlinearsegment
  \!xS=\!xE \!yS=\!yE
  \ignorespaces}
\def\!linearsolid{%
  \!npoints=\!arclength
  \!countA=\plotsymbolspacing
  \divide\!npoints by \!countA
  \ifnum \!npoints<1 
    \!npoints=1 
  \fi
  \divide\!xdiff by \!npoints
  \divide\!ydiff by \!npoints
  \!xpos=\!xS \!ypos=\!yS
  \loop\ifnum\!npoints>-1
    \!plotifinbounds
    \advance \!xpos by \!xdiff
    \advance \!ypos by \!ydiff
    \advance \!npoints by -1
  \repeat
  \ignorespaces}
\def\!lineardashed{%
  \ifdim\!distacross>\!arclength
    \advance \!distacross by -\!arclength  
  \else
    \loop\ifdim\!distacross<\!arclength
      \!divide\!distacross\!arclength\!dimenA
      \!removept\!dimenA\!t
      \!xpos=\!t\!xdiff \advance \!xpos by \!xS
      \!ypos=\!t\!ydiff \advance \!ypos by \!yS
      \!plotifinbounds
      \advance\!distacross by \plotsymbolspacing
      \!advancedashing
    \repeat  
    \advance \!distacross by -\!arclength
  \fi
  \ignorespaces}
\def\!!advancedashing{%
  \advance\!downlength by -\plotsymbolspacing
  \ifdim \!downlength>\!zpt
  \else
    \advance\!distacross by \!downlength
    \!getnextvalueof\!uplength\from\!!UDlist
    \advance\!distacross by \!uplength
    \!getnextvalueof\!downlength\from\!!UDlist
  \fi}
\def\inboundscheckoff{%
  \def\!plotifinbounds{\!plot(\!xpos,\!ypos)}%
  \def\!initinboundscheck{\relax}\ignorespaces}
 
\inboundscheckoff
\def\!!plotifinbounds{%
  \ifdim \!xpos<\!checkleft
  \else
    \ifdim \!xpos>\!checkright
    \else
      \ifdim \!ypos<\!checkbot
      \else
         \ifdim \!ypos>\!checktop
         \else
           \!plot(\!xpos,\!ypos)
         \fi 
      \fi
    \fi
  \fi}
\def\!!initinboundscheck{%
  \!checkleft=\!arealloc     \advance\!checkleft by \!xorigin
  \!checkright=\!arearloc    \advance\!checkright by \!xorigin
  \!checkbot=\!areabloc      \advance\!checkbot by \!yorigin
  \!checktop=\!areatloc      \advance\!checktop by \!yorigin}
\def\!logten#1#2{%
  \expandafter\!!logten#1\!nil
  \!removept\!dimenF#2%
  \ignorespaces}
\def\!!logten#1#2\!nil{%
  \if -#1%
    \!dimenF=\!zpt
    \def\!next{\ignorespaces}%
  \else
    \if +#1%
      \def\!next{\!!logten#2\!nil}%
    \else
      \if .#1%
        \def\!next{\!!logten0.#2\!nil}%
      \else
        \def\!next{\!!!logten#1#2..\!nil}%
      \fi
    \fi
  \fi
  \!next}
\def\!!!logten#1#2.#3.#4\!nil{%
  \!dimenF=1pt 
  \if 0#1%
    \!!logshift#3pt 
  \else 
    \!logshift#2/
    \!dimenE=#1.#2#3pt 
  \fi 
  \ifdim \!dimenE<\!rootten
    \multiply \!dimenE 10 
    \advance  \!dimenF -1pt
  \fi
  \!dimenG=\!dimenE
    \advance\!dimenG 10pt
  \advance\!dimenE -10pt 
  \multiply\!dimenE 10 
  \!divide\!dimenE\!dimenG\!dimenE
  \!removept\!dimenE\!t
  \!dimenG=\!t\!dimenE
  \!removept\!dimenG\!tt
  \!dimenH=\!tt\!tenAe
    \divide\!dimenH 100
  \advance\!dimenH \!tenAc
  \!dimenH=\!tt\!dimenH
    \divide\!dimenH 100   
  \advance\!dimenH \!tenAa
  \!dimenH=\!t\!dimenH
    \divide\!dimenH 100 
  \advance\!dimenF \!dimenH}
\def\!logshift#1{%
  \if #1/%
    \def\!next{\ignorespaces}%
  \else
    \advance\!dimenF 1pt 
    \def\!next{\!logshift}%
  \fi 
  \!next}
 \def\!!logshift#1{%
   \advance\!dimenF -1pt
   \if 0#1%
     \def\!next{\!!logshift}%
   \else
     \if p#1%
       \!dimenF=1pt
       \def\!next{\!dimenE=1p}%
     \else
       \def\!next{\!dimenE=#1.}%
     \fi
   \fi
   \!next}
\def\beginpicture{%
  \setbox\!picbox=\hbox\bgroup%
  \!xleft=\maxdimen  
  \!xright=-\maxdimen
  \!ybot=\maxdimen
  \!ytop=-\maxdimen}
\def\endpicture{%
  \ifdim\!xleft=\maxdimen
    \!xleft=\!zpt \!xright=\!zpt \!ybot=\!zpt \!ytop=\!zpt 
  \fi
  \global\!Xleft=\!xleft \global\!Xright=\!xright
  \global\!Ybot=\!ybot \global\!Ytop=\!ytop
  \egroup%
  \ht\!picbox=\!Ytop  \dp\!picbox=-\!Ybot
  \ifdim\!Ybot>\!zpt
  \else 
    \ifdim\!Ytop<\!zpt
      \!Ybot=\!Ytop
    \else
      \!Ybot=\!zpt
    \fi
  \fi
  \hbox{\kern-\!Xleft\lower\!Ybot\box\!picbox\kern\!Xright}}
\def\endpicturesave <#1,#2>{%
  \endpicture \global #1=\!Xleft \global #2=\!Ybot \ignorespaces}
\def\setcoordinatesystem{%
  \!ifnextchar{u}{\!getlengths }
    {\!getlengths units <\!xunit,\!yunit>}}
\def\!getlengths units <#1,#2>{%
  \!xunit=#1\relax
  \!yunit=#2\relax
  \!ifcoordmode 
    \let\!SCnext=\!SCccheckforRP
  \else
    \let\!SCnext=\!SCdcheckforRP
  \fi
  \!SCnext}
\def\!SCccheckforRP{%
  \!ifnextchar{p}{\!cgetreference }
    {\!cgetreference point at {\!xref} {\!yref} }}
\def\!cgetreference point at #1 #2 {%
  \edef\!xref{#1}\edef\!yref{#2}%
  \!xorigin=\!xref\!xunit  \!yorigin=\!yref\!yunit  
  \!initinboundscheck 
  \ignorespaces}
\def\!SCdcheckforRP{%
  \!ifnextchar{p}{\!dgetreference}%
    {\ignorespaces}}
\def\!dgetreference point at #1 #2 {%
  \!xorigin=#1\relax  \!yorigin=#2\relax
  \ignorespaces}
\long\def\put#1#2 at #3 #4 {%
  \!setputobject{#1}{#2}%
  \!xpos=\!M{#3}\!xunit  \!ypos=\!M{#4}\!yunit  
  \!rotateaboutpivot\!xpos\!ypos%
  \advance\!xpos -\!xorigin  \advance\!xpos -\!xshift
  \advance\!ypos -\!yorigin  \advance\!ypos -\!yshift
  \kern\!xpos\raise\!ypos\box\!putobject\kern-\!xpos%
  \!doaccounting\ignorespaces}
\long\def\multiput #1#2 at {%
  \!setputobject{#1}{#2}%
  \!ifnextchar"{\!putfromfile}{\!multiput}}
\def\!putfromfile"#1"{%
  \expandafter\!multiput \input #1 /}
\def\!multiput{%
  \futurelet\!nextchar\!!multiput}
\def\!!multiput{%
  \if *\!nextchar
    \def\!nextput{\!alsoby}%
  \else
    \if /\!nextchar
      \def\!nextput{\!finishmultiput}%
    \else
      \def\!nextput{\!alsoat}%
    \fi
  \fi
  \!nextput}
\def\!finishmultiput/{%
  \setbox\!putobject=\hbox{}%
  \ignorespaces}
\def\!alsoat#1 #2 {%
  \!xpos=\!M{#1}\!xunit  \!ypos=\!M{#2}\!yunit  
  \!rotateaboutpivot\!xpos\!ypos%
  \advance\!xpos -\!xorigin  \advance\!xpos -\!xshift
  \advance\!ypos -\!yorigin  \advance\!ypos -\!yshift
  \kern\!xpos\raise\!ypos\copy\!putobject\kern-\!xpos%
  \!doaccounting
  \!multiput}
\def\!alsoby*#1 #2 #3 {%
  \!dxpos=\!M{#2}\!xunit \!dypos=\!M{#3}\!yunit 
  \!rotateonly\!dxpos\!dypos
  \!ntemp=#1%
  \!!loop\ifnum\!ntemp>0
    \advance\!xpos by \!dxpos  \advance\!ypos by \!dypos
    \kern\!xpos\raise\!ypos\copy\!putobject\kern-\!xpos%
    \advance\!ntemp by -1
  \repeat
  \!doaccounting 
  \!multiput}
\def\accountingon{\def\!doaccounting{\!!doaccounting}\ignorespaces}

\accountingon
\def\!!doaccounting{%
  \!xtemp=\!xpos  
  \!ytemp=\!ypos
  \ifdim\!xtemp<\!xleft 
     \!xleft=\!xtemp 
  \fi
  \advance\!xtemp by  \!wd 
  \ifdim\!xright<\!xtemp 
    \!xright=\!xtemp
  \fi
  \advance\!ytemp by -\!dp
  \ifdim\!ytemp<\!ybot  
    \!ybot=\!ytemp
  \fi
  \advance\!ytemp by  \!dp
  \advance\!ytemp by  \!ht 
  \ifdim\!ytemp>\!ytop  
    \!ytop=\!ytemp  
  \fi}
\long\def\!setputobject#1#2{%
  \setbox\!putobject=\hbox{#1}%
  \!ht=\ht\!putobject  \!dp=\dp\!putobject  \!wd=\wd\!putobject
  \wd\!putobject=\!zpt
  \!xshift=.5\!wd   \!yshift=.5\!ht   \advance\!yshift by -.5\!dp
  \edef\!putorientation{#2}%
  \expandafter\!SPOreadA\!putorientation[]\!nil%
  \expandafter\!SPOreadB\!putorientation<\!zpt,\!zpt>\!nil\ignorespaces}
\def\!SPOreadA#1[#2]#3\!nil{\!etfor\!orientation:=#2\do\!SPOreviseshift}
\def\!SPOreadB#1<#2,#3>#4\!nil{\advance\!xshift by -#2\advance\!yshift by -#3}
\def\!SPOreviseshift{%
  \if l\!orientation 
    \!xshift=\!zpt
  \else 
    \if r\!orientation 
      \!xshift=\!wd
    \else 
      \if b\!orientation
        \!yshift=-\!dp
      \else 
        \if B\!orientation 
          \!yshift=\!zpt
        \else 
          \if t\!orientation 
            \!yshift=\!ht
          \fi 
        \fi
      \fi
    \fi
  \fi}
\long\def\!dimenput#1#2(#3,#4){%
  \!setputobject{#1}{#2}%
  \!xpos=#3\advance\!xpos by -\!xshift
  \!ypos=#4\advance\!ypos by -\!yshift
  \kern\!xpos\raise\!ypos\box\!putobject\kern-\!xpos%
  \!doaccounting\ignorespaces}
\def\!setdimenmode{%
  \let\!M=\!M!!\ignorespaces}
\def\!setcoordmode{%
  \let\!M=\!M!\ignorespaces}
\def\!ifcoordmode{%
  \ifx \!M \!M!}
\def\!ifdimenmode{%
  \ifx \!M \!M!!}
\def\!M!#1#2{#1#2} 
\def\!M!!#1#2{#1}
\!setcoordmode
\let\setdimensionmode=\!setdimenmode
\let\setcoordinatemode=\!setcoordmode

\def\!stack[#1]{%
  \let\!lglue=\hfill \let\!rglue=\hfill
  \expandafter\let\csname !#1glue\endcsname=\relax
  \!ifnextchar<{\!!stack}{\!!stack<\stackleading>}}
\def\!!stack<#1>#2{%
  \vbox{\def\!valueslist{}\!ecfor\!value:=#2\do{%
    \expandafter\!rightappend\!value\withCS{\\}\to\!valueslist}%
    \!lop\!valueslist\to\!value
    \let\\=\cr\lineskiplimit=\maxdimen\lineskip=#1%
    \baselineskip=-1000pt\halign{\!lglue##\!rglue\cr \!value\!valueslist\cr}}%
  \ignorespaces}

\def\!lines[#1]#2{%
  \let\!lglue=\hfill \let\!rglue=\hfill
  \expandafter\let\csname !#1glue\endcsname=\relax
  \vbox{\halign{\!lglue##\!rglue\cr #2\crcr}}%
  \ignorespaces}

\def\!Lines[#1]#2{%
  \let\!lglue=\hfill \let\!rglue=\hfill
  \expandafter\let\csname !#1glue\endcsname=\relax
  \vtop{\halign{\!lglue##\!rglue\cr #2\crcr}}%
  \ignorespaces}
\def\setplotsymbol(#1#2){%
  \!setputobject{#1}{#2}
  \setbox\!plotsymbol=\box\!putobject%
  \!plotsymbolxshift=\!xshift 
  \!plotsymbolyshift=\!yshift 
  \ignorespaces}
\setplotsymbol({\fiverm .})
\def\!!plot(#1,#2){%
  \!dimenA=-\!plotxorigin \advance \!dimenA by #1
  \!dimenB=-\!plotyorigin \advance \!dimenB by #2
  \kern\!dimenA\raise\!dimenB\copy\!plotsymbol\kern-\!dimenA%
  \ignorespaces}
\def\!!!plot(#1,#2){%
  \!dimenA=-\!plotxorigin \advance \!dimenA by #1
  \!dimenB=-\!plotyorigin \advance \!dimenB by #2
  \kern\!dimenA\raise\!dimenB\copy\!plotsymbol\kern-\!dimenA%
  \!countE=\!dimenA
  \!countF=\!dimenB
  \immediate\write\!replotfile{\the\!countE,\the\!countF.}%
  \ignorespaces}
\def\savelinesandcurves on "#1" {%
  \immediate\closeout\!replotfile
  \immediate\openout\!replotfile=#1%
  \let\!plot=\!!!plot}
\def\dontsavelinesandcurves {%
  \let\!plot=\!!plot}
\dontsavelinesandcurves
{\catcode`\%=11\xdef\!Commentsignal{
\def\writesavefile#1 {%
  \immediate\write\!replotfile{\!Commentsignal #1}%
  \ignorespaces}
\def\replot"#1" {%
  \expandafter\!replot\input #1 /}
\def\!replot#1,#2. {%
  \!dimenA=#1sp
  \kern\!dimenA\raise#2sp\copy\!plotsymbol\kern-\!dimenA
  \futurelet\!nextchar\!!replot}
\def\!!replot{%
  \if /\!nextchar 
    \def\!next{\!finish}%
  \else
    \def\!next{\!replot}%
  \fi
  \!next}
\def\!Pythag#1#2#3{%
  \!dimenE=#1\relax                                     
  \ifdim\!dimenE<\!zpt 
    \!dimenE=-\!dimenE 
  \fi
  \!dimenF=#2\relax
  \ifdim\!dimenF<\!zpt 
    \!dimenF=-\!dimenF 
  \fi
  \advance \!dimenF by \!dimenE
  \ifdim\!dimenF=\!zpt 
    \!dimenG=\!zpt
  \else 
    \!divide{8\!dimenE}\!dimenF\!dimenE
    \advance\!dimenE by -4pt
      \!dimenE=2\!dimenE
    \!removept\!dimenE\!!t
    \!dimenE=\!!t\!dimenE
    \advance\!dimenE by 64pt
    \divide \!dimenE by 2
    \!dimenH=7pt
    \!!Pythag\!!Pythag\!!Pythag
    \!removept\!dimenH\!!t
    \!dimenG=\!!t\!dimenF
    \divide\!dimenG by 8
  \fi
  #3=\!dimenG
  \ignorespaces}
\def\!!Pythag{
  \!divide\!dimenE\!dimenH\!dimenI
  \advance\!dimenH by \!dimenI
    \divide\!dimenH by 2}
\def\placehypotenuse for <#1> and <#2> in <#3> {%
  \!Pythag{#1}{#2}{#3}}
\def\!qjoin (#1,#2) (#3,#4){%
  \advance\!intervalno by 1
  \!ifcoordmode
    \edef\!xmidpt{#1}\edef\!ymidpt{#2}%
  \else
    \!dimenA=#1\relax \edef\!xmidpt{\the\!dimenA}%
    \!dimenA=#2\relax \edef\!xmidpt{\the\!dimenA}%
  \fi
  \!xM=\!M{#1}\!xunit  \!yM=\!M{#2}\!yunit   \!rotateaboutpivot\!xM\!yM
  \!xE=\!M{#3}\!xunit  \!yE=\!M{#4}\!yunit   \!rotateaboutpivot\!xE\!yE
  \!dimenA=\!xM  \advance \!dimenA by -\!xS
  \!dimenB=\!xE  \advance \!dimenB by -\!xM
  \!xB=3\!dimenA \advance \!xB by -\!dimenB
  \!xC=2\!dimenB \advance \!xC by -2\!dimenA
  \!dimenA=\!yM  \advance \!dimenA by -\!yS%
  \!dimenB=\!yE  \advance \!dimenB by -\!yM%
  \!yB=3\!dimenA \advance \!yB by -\!dimenB%
  \!yC=2\!dimenB \advance \!yC by -2\!dimenA%
  \!xprime=\!xB  \!yprime=\!yB
  \!dxprime=.5\!xC  \!dyprime=.5\!yC
  \!getf \!midarclength=\!dimenA
  \!getf \advance \!midarclength by 4\!dimenA
  \!getf \advance \!midarclength by \!dimenA
  \divide \!midarclength by 12
  \!arclength=\!dimenA
  \!getf \advance \!arclength by 4\!dimenA
  \!getf \advance \!arclength by \!dimenA
  \divide \!arclength by 12
  \advance \!arclength by \!midarclength
  \global\advance \totalarclength by \!arclength
  \ifdim\!distacross>\!arclength 
    \advance \!distacross by -\!arclength
  \else
    \!initinverseinterp
    \loop\ifdim\!distacross<\!arclength
      \!inverseinterp
      \!xpos=\!t\!xC \advance\!xpos by \!xB
        \!xpos=\!t\!xpos \advance \!xpos by \!xS
      \!ypos=\!t\!yC \advance\!ypos by \!yB
        \!ypos=\!t\!ypos \advance \!ypos by \!yS
      \!plotifinbounds
      \advance\!distacross \plotsymbolspacing
      \!advancedashing
    \repeat  
    \advance \!distacross by -\!arclength
  \fi
  \!xS=\!xE
  \!yS=\!yE
  \ignorespaces}
\def\!getf{\!Pythag\!xprime\!yprime\!dimenA%
  \advance\!xprime by \!dxprime
  \advance\!yprime by \!dyprime}
\def\!initinverseinterp{%
  \ifdim\!arclength>\!zpt
    \!divide{8\!midarclength}\!arclength\!dimenE
    \ifdim\!dimenE<\!wmin \!setinverselinear
    \else 
      \ifdim\!dimenE>\!wmax \!setinverselinear
      \else
        \def\!inverseinterp{\!inversequad}\ignorespaces
         \!removept\!dimenE\!Ew
         \!dimenF=-\!Ew\!dimenE
         \advance\!dimenF by 32pt
         \!dimenG=8pt 
         \advance\!dimenG by -\!dimenE
         \!dimenG=\!Ew\!dimenG
         \!divide\!dimenF\!dimenG\!beta
         \!gamma=1pt
         \advance \!gamma by -\!beta
      \fi
    \fi
  \fi
  \ignorespaces}
\def\!inversequad{%
  \!divide\!distacross\!arclength\!dimenG
  \!removept\!dimenG\!v
  \!dimenG=\!v\!gamma
  \advance\!dimenG by \!beta
  \!dimenG=\!v\!dimenG
  \!removept\!dimenG\!t}
\def\!setinverselinear{%
  \def\!inverseinterp{\!inverselinear}%
  \divide\!dimenE by 8 \!removept\!dimenE\!t
  \!countC=\!intervalno \multiply \!countC 2
  \!countB=\!countC     \advance \!countB -1
  \!countA=\!countB     \advance \!countA -1
  \wlog{\the\!countB th point (\!xmidpt,\!ymidpt) being plotted 
    doesn't lie in the}%
  \wlog{ middle third of the arc between the \the\!countA th 
    and \the\!countC th points:}%
  \wlog{ [arc length \the\!countA\space to \the\!countB]/[arc length 
    \the \!countA\space to \the\!countC]=\!t.}%
  \ignorespaces}
\def\!inverselinear{%
  \!divide\!distacross\!arclength\!dimenG
  \!removept\!dimenG\!t}
\def\startrotation{%
  \let\!rotateaboutpivot=\!!rotateaboutpivot
  \let\!rotateonly=\!!rotateonly
  \!ifnextchar{b}{\!getsincos }%
    {\!getsincos by {\!cosrotationangle} {\!sinrotationangle} }}
\def\!getsincos by #1 #2 {%
  \edef\!cosrotationangle{#1}%
  \edef\!sinrotationangle{#2}%
  \!ifcoordmode 
    \let\!ROnext=\!ccheckforpivot
  \else
    \let\!ROnext=\!dcheckforpivot
  \fi
  \!ROnext}
\def\!ccheckforpivot{%
  \!ifnextchar{a}{\!cgetpivot}%
    {\!cgetpivot about {\!xpivotcoord} {\!ypivotcoord} }}
\def\!cgetpivot about #1 #2 {%
  \edef\!xpivotcoord{#1}%
  \edef\!ypivotcoord{#2}%
  \!xpivot=#1\!xunit  \!ypivot=#2\!yunit
  \ignorespaces}
\def\!dcheckforpivot{%
  \!ifnextchar{a}{\!dgetpivot}{\ignorespaces}}
\def\!dgetpivot about #1 #2 {%
  \!xpivot=#1\relax  \!ypivot=#2\relax
  \ignorespaces}
\def\stoprotation{%
  \let\!rotateaboutpivot=\!!!rotateaboutpivot
  \let\!rotateonly=\!!!rotateonly
  \ignorespaces}
\def\!!rotateaboutpivot#1#2{%
  \!dimenA=#1\relax  \advance\!dimenA -\!xpivot
  \!dimenB=#2\relax  \advance\!dimenB -\!ypivot
  \!dimenC=\!cosrotationangle\!dimenA
    \advance \!dimenC -\!sinrotationangle\!dimenB
  \!dimenD=\!cosrotationangle\!dimenB
    \advance \!dimenD  \!sinrotationangle\!dimenA
  \advance\!dimenC \!xpivot  \advance\!dimenD \!ypivot
  #1=\!dimenC  #2=\!dimenD
  \ignorespaces}
\def\!!rotateonly#1#2{%
  \!dimenA=#1\relax  \!dimenB=#2\relax 
  \!dimenC=\!cosrotationangle\!dimenA
    \advance \!dimenC -\!rotsign\!sinrotationangle\!dimenB
  \!dimenD=\!cosrotationangle\!dimenB
    \advance \!dimenD  \!rotsign\!sinrotationangle\!dimenA
  #1=\!dimenC  #2=\!dimenD
  \ignorespaces}
\def\!rotsign{}
\def\!!!rotateaboutpivot#1#2{\relax}
\def\!!!rotateonly#1#2{\relax}
\stoprotation
\def\!reverserotateonly#1#2{%
  \def\!rotsign{-}%
  \!rotateonly{#1}{#2}%
  \def\!rotsign{}%
  \ignorespaces}

\def\!getspan span <#1>{%
  \!dshade=#1\relax
  \!ifcoordmode 
    \let\!GRnext=\!GRccheckforAP
  \else
    \let\!GRnext=\!GRdcheckforAP
  \fi
  \!GRnext}
\def\!GRccheckforAP{%
  \!ifnextchar{p}{\!cgetanchor }
    {\!cgetanchor point at {\!xshadesave} {\!yshadesave} }}
\def\!cgetanchor point at #1 #2 {%
  \edef\!xshadesave{#1}\edef\!yshadesave{#2}%
  \!xshade=\!xshadesave\!xunit  \!yshade=\!yshadesave\!yunit
  \ignorespaces}
\def\!GRdcheckforAP{%
  \!ifnextchar{p}{\!dgetanchor}%
    {\ignorespaces}}
\def\!dgetanchor point at #1 #2 {%
  \!xshade=#1\relax  \!yshade=#2\relax
  \ignorespaces}
\def\setshadesymbol{%
  \!ifnextchar<{\!setshadesymbol}{\!setshadesymbol<,,,> }}
\def\!setshadesymbol <#1,#2,#3,#4> (#5#6){%
  \!setputobject{#5}{#6}%
  \setbox\!shadesymbol=\box\!putobject%
  \!shadesymbolxshift=\!xshift \!shadesymbolyshift=\!yshift
  \!dimenA=\!xshift \advance\!dimenA \!smidge
  \!override\!dimenA{#1}\!lshrinkage%
  \!dimenA=\!wd \advance \!dimenA -\!xshift
    \advance\!dimenA \!smidge
    \!override\!dimenA{#2}\!rshrinkage
  \!dimenA=\!dp \advance \!dimenA \!yshift
    \advance\!dimenA \!smidge
    \!override\!dimenA{#3}\!bshrinkage
  \!dimenA=\!ht \advance \!dimenA -\!yshift
    \advance\!dimenA \!smidge
    \!override\!dimenA{#4}\!tshrinkage
  \ignorespaces}
\def\!smidge{-.2pt}%
\def\!override#1#2#3{%
  \edef\!!override{#2}%
  \ifx \!!override\empty
    #3=#1\relax
  \else
    \if z\!!override
      #3=\!zpt
    \else
      \ifx \!!override\!blankz
        #3=\!zpt
      \else
        #3=#2\relax
      \fi
    \fi
  \fi
  \ignorespaces}
\def\!blankz{ z}
\setshadesymbol ({\fiverm .})
\def\!startvshade#1(#2,#3,#4){%
  \let\!!xunit=\!xunit%
  \let\!!yunit=\!yunit%
  \let\!!xshade=\!xshade%
  \let\!!yshade=\!yshade%
  \def\!getshrinkages{\!vgetshrinkages}%
  \let\!setshadelocation=\!vsetshadelocation%
  \!xS=\!M{#2}\!!xunit
  \!ybS=\!M{#3}\!!yunit
  \!ytS=\!M{#4}\!!yunit
  \!shadexorigin=\!xorigin  \advance \!shadexorigin \!shadesymbolxshift
  \!shadeyorigin=\!yorigin  \advance \!shadeyorigin \!shadesymbolyshift
  \ignorespaces}
\def\!starthshade#1(#2,#3,#4){%
  \let\!!xunit=\!yunit%
  \let\!!yunit=\!xunit%
  \let\!!xshade=\!yshade%
  \let\!!yshade=\!xshade%
  \def\!getshrinkages{\!hgetshrinkages}%
  \let\!setshadelocation=\!hsetshadelocation%
  \!xS=\!M{#2}\!!xunit
  \!ybS=\!M{#3}\!!yunit
  \!ytS=\!M{#4}\!!yunit
  \!shadexorigin=\!xorigin  \advance \!shadexorigin \!shadesymbolxshift
  \!shadeyorigin=\!yorigin  \advance \!shadeyorigin \!shadesymbolyshift
  \ignorespaces}
\def\!lattice#1#2#3#4#5{%
  \!dimenA=#1
  \!dimenB=#2
  \!countB=\!dimenB
  \!dimenC=#3
  \advance\!dimenC -\!dimenA
  \!countA=\!dimenC
  \divide\!countA \!countB
  \ifdim\!dimenC>\!zpt
    \!dimenD=\!countA\!dimenB
    \ifdim\!dimenD<\!dimenC
      \advance\!countA 1 
    \fi
  \fi
  \!dimenC=\!countA\!dimenB
    \advance\!dimenC \!dimenA
  #4=\!countA
  #5=\!dimenC
  \ignorespaces}
\def\!qshade#1(#2,#3,#4)#5(#6,#7,#8){%
  \!xM=\!M{#2}\!!xunit
  \!ybM=\!M{#3}\!!yunit
  \!ytM=\!M{#4}\!!yunit
  \!xE=\!M{#6}\!!xunit
  \!ybE=\!M{#7}\!!yunit
  \!ytE=\!M{#8}\!!yunit
  \!getcoeffs\!xS\!ybS\!xM\!ybM\!xE\!ybE\!ybB\!ybC
  \!getcoeffs\!xS\!ytS\!xM\!ytM\!xE\!ytE\!ytB\!ytC
  \def\!getylimits{\!qgetylimits}%
  \!shade{#1}\ignorespaces}
\def\!lshade#1(#2,#3,#4){%
  \!xE=\!M{#2}\!!xunit
  \!ybE=\!M{#3}\!!yunit
  \!ytE=\!M{#4}\!!yunit
  \!dimenE=\!xE  \advance \!dimenE -\!xS
  \!dimenC=\!ytE \advance \!dimenC -\!ytS
  \!divide\!dimenC\!dimenE\!ytB
  \!dimenC=\!ybE \advance \!dimenC -\!ybS
  \!divide\!dimenC\!dimenE\!ybB
  \def\!getylimits{\!lgetylimits}%
  \!shade{#1}\ignorespaces}
\def\!getcoeffs#1#2#3#4#5#6#7#8{%
  \!dimenC=#4\advance \!dimenC -#2
  \!dimenE=#3\advance \!dimenE -#1
  \!divide\!dimenC\!dimenE\!dimenF
  \!dimenC=#6\advance \!dimenC -#4
  \!dimenH=#5\advance \!dimenH -#3
  \!divide\!dimenC\!dimenH\!dimenG
  \advance\!dimenG -\!dimenF
  \advance \!dimenH \!dimenE
  \!divide\!dimenG\!dimenH#8
  \!removept#8\!t
  #7=-\!t\!dimenE
  \advance #7\!dimenF
  \ignorespaces}
\def\!shade#1{%
  \!getshrinkages#1<,,,>\!nil
  \advance \!dimenE \!xS
  \!lattice\!!xshade\!dshade\!dimenE
    \!parity\!xpos
  \!dimenF=-\!dimenF
    \advance\!dimenF \!xE
  \!loop\!not{\ifdim\!xpos>\!dimenF}
    \!shadecolumn%
    \advance\!xpos \!dshade
    \advance\!parity 1
  \repeat
  \!xS=\!xE
  \!ybS=\!ybE
  \!ytS=\!ytE
  \ignorespaces}
\def\!vgetshrinkages#1<#2,#3,#4,#5>#6\!nil{%
  \!override\!lshrinkage{#2}\!dimenE
  \!override\!rshrinkage{#3}\!dimenF
  \!override\!bshrinkage{#4}\!dimenG
  \!override\!tshrinkage{#5}\!dimenH
  \ignorespaces}
\def\!hgetshrinkages#1<#2,#3,#4,#5>#6\!nil{%
  \!override\!lshrinkage{#2}\!dimenG
  \!override\!rshrinkage{#3}\!dimenH
  \!override\!bshrinkage{#4}\!dimenE
  \!override\!tshrinkage{#5}\!dimenF
  \ignorespaces}
\def\!shadecolumn{%
  \!dxpos=\!xpos
  \advance\!dxpos -\!xS
  \!removept\!dxpos\!dx
  \!getylimits
  \advance\!ytpos -\!dimenH
  \advance\!ybpos \!dimenG
  \!yloc=\!!yshade
  \ifodd\!parity 
     \advance\!yloc \!dshade
  \fi
  \!lattice\!yloc{2\!dshade}\!ybpos%
    \!countA\!ypos
  \!dimenA=-\!shadexorigin \advance \!dimenA \!xpos
  \loop\!not{\ifdim\!ypos>\!ytpos}
    \!setshadelocation
    \!rotateaboutpivot\!xloc\!yloc%
    \!dimenA=-\!shadexorigin \advance \!dimenA \!xloc
    \!dimenB=-\!shadeyorigin \advance \!dimenB \!yloc
    \kern\!dimenA \raise\!dimenB\copy\!shadesymbol \kern-\!dimenA
    \advance\!ypos 2\!dshade
  \repeat
  \ignorespaces}
\def\!qgetylimits{%
  \!dimenA=\!dx\!ytC              
  \advance\!dimenA \!ytB
  \!ytpos=\!dx\!dimenA
  \advance\!ytpos \!ytS
  \!dimenA=\!dx\!ybC              
  \advance\!dimenA \!ybB
  \!ybpos=\!dx\!dimenA
  \advance\!ybpos \!ybS}
\def\!lgetylimits{%
  \!ytpos=\!dx\!ytB
  \advance\!ytpos \!ytS
  \!ybpos=\!dx\!ybB
  \advance\!ybpos \!ybS}
\def\!vsetshadelocation{
  \!xloc=\!xpos
  \!yloc=\!ypos}
\def\!hsetshadelocation{
  \!xloc=\!ypos
  \!yloc=\!xpos}
\def\!axisticks {%
  \def\!nextkeyword##1 {%
    \expandafter\ifx\csname !ticks##1\endcsname \relax
      \def\!next{\!fixkeyword{##1}}%
    \else
      \def\!next{\csname !ticks##1\endcsname}%
    \fi
    \!next}%
  \!axissetup
    \def\!axissetup{\relax}%
  \edef\!ticksinoutsign{\!ticksinoutSign}%
  \!ticklength=\longticklength
  \!tickwidth=\linethickness
  \!gridlinestatus
  \!setticktransform
  \!maketick
  \!tickcase=0
  \def\!LTlist{}%
  \!nextkeyword}
\def\ticksout{%
  \def\!ticksinoutSign{+}}

\ticksout

\def\nogridlines{%
  \def\!gridlinestatus{\!gridlinestoofalse}}
\nogridlines
\def\loggedticks{%
  \def\!setticktransform{\let\!ticktransform=\!logten}}
\def\unloggedticks{%
  \def\!setticktransform{\let\!ticktransform=\!donothing}}
\def\!donothing#1#2{\def#2{#1}}
\unloggedticks
\expandafter\def\csname !ticks/\endcsname{%
  \!not {\ifx \!LTlist\empty}
    \!placetickvalues
  \fi
  \def\!tickvalueslist{}%
  \def\!LTlist{}%
  \expandafter\csname !axis/\endcsname}
\def\!maketick{%
  \setbox\!boxA=\hbox{%
    \beginpicture
      \!setdimenmode
      \setcoordinatesystem point at {\!zpt} {\!zpt}   
      \linethickness=\!tickwidth
      \ifdim\!ticklength>\!zpt
        \putrule from {\!zpt} {\!zpt} to
          {\!ticksinoutsign\!tickxsign\!ticklength}
          {\!ticksinoutsign\!tickysign\!ticklength}
      \fi
      \if!gridlinestoo
        \putrule from {\!zpt} {\!zpt} to
          {-\!tickxsign\!xaxislength} {-\!tickysign\!yaxislength}
      \fi
    \endpicturesave <\!Xsave,\!Ysave>}%
    \wd\!boxA=\!zpt}
\def\!ticksin{%
  \def\!ticksinoutsign{-}%
  \!maketick
  \!nextkeyword}
\def\!ticksout{%
  \def\!ticksinoutsign{+}%
  \!maketick
  \!nextkeyword}
\def\!tickslength<#1> {%
  \!ticklength=#1\relax
  \!maketick
  \!nextkeyword}
\def\!tickslong{%
  \!tickslength<\longticklength> }
\def\!ticksshort{%
  \!tickslength<\shortticklength> }
\def\!tickswidth<#1> {%
  \!tickwidth=#1\relax
  \!maketick
  \!nextkeyword}
\def\!ticksandacross{%
  \!gridlinestootrue
  \!maketick
  \!nextkeyword}
\def\!ticksbutnotacross{%
  \!gridlinestoofalse
  \!maketick
  \!nextkeyword}
\def\!tickslogged{%
  \let\!ticktransform=\!logten
  \!nextkeyword}
\def\!ticksunlogged{%
  \let\!ticktransform=\!donothing
  \!nextkeyword}
\def\!ticksunlabeled{%
  \!tickcase=0
  \!nextkeyword}
\def\!ticksnumbered{%
  \!tickcase=1
  \!nextkeyword}
\def\!tickswithvalues#1/ {%
  \edef\!tickvalueslist{#1! /}%
  \!tickcase=2
  \!nextkeyword}
\def\!ticksquantity#1 {%
  \ifnum #1>1
    \!updatetickoffset
    \!countA=#1\relax
    \advance \!countA -1
    \!ticklocationincr=\!axisLength
      \divide \!ticklocationincr \!countA
    \!ticklocation=\!axisstart
    \loop \!not{\ifdim \!ticklocation>\!axisend}
      \!placetick\!ticklocation
      \ifcase\!tickcase
          \relax 
        \or
          \relax 
        \or
          \expandafter\!gettickvaluefrom\!tickvalueslist
          \edef\!tickfield{{\the\!ticklocation}{\!value}}%
          \expandafter\!listaddon\expandafter{\!tickfield}\!LTlist%
      \fi
      \advance \!ticklocation \!ticklocationincr
    \repeat
  \fi
  \!nextkeyword}
\def\!ticksat#1 {%
  \!updatetickoffset
  \edef\!Loc{#1}%
  \if /\!Loc
    \def\next{\!nextkeyword}%
  \else
    \!ticksincommon
    \def\next{\!ticksat}%
  \fi
  \next}    
\def\!ticksfrom#1 to #2 by #3 {%
  \!updatetickoffset
  \edef\!arg{#3}%
  \expandafter\!separate\!arg\!nil
  \!scalefactor=1
  \expandafter\!countfigures\!arg/
  \edef\!arg{#1}%
  \!scaleup\!arg by\!scalefactor to\!countE
  \edef\!arg{#2}%
  \!scaleup\!arg by\!scalefactor to\!countF
  \edef\!arg{#3}%
  \!scaleup\!arg by\!scalefactor to\!countG
  \loop \!not{\ifnum\!countE>\!countF}
    \ifnum\!scalefactor=1
      \edef\!Loc{\the\!countE}%
    \else
      \!scaledown\!countE by\!scalefactor to\!Loc
    \fi
    \!ticksincommon
    \advance \!countE \!countG
  \repeat
  \!nextkeyword}
\def\!updatetickoffset{%
  \!dimenA=\!ticksinoutsign\!ticklength
  \ifdim \!dimenA>\!offset
    \!offset=\!dimenA
  \fi}
\def\!placetick#1{%
  \if!xswitch
    \!xpos=#1\relax
    \!ypos=\!axisylevel
  \else
    \!xpos=\!axisxlevel
    \!ypos=#1\relax
  \fi
  \advance\!xpos \!Xsave
  \advance\!ypos \!Ysave
  \kern\!xpos\raise\!ypos\copy\!boxA\kern-\!xpos
  \ignorespaces}
\def\!gettickvaluefrom#1 #2 /{%
  \edef\!value{#1}%
  \edef\!tickvalueslist{#2 /}%
  \ifx \!tickvalueslist\!endtickvaluelist
    \!tickcase=0
  \fi}
\def\!endtickvaluelist{! /}
\def\!ticksincommon{%
  \!ticktransform\!Loc\!t
  \!ticklocation=\!t\!!unit
  \advance\!ticklocation -\!!origin
  \!placetick\!ticklocation
  \ifcase\!tickcase
    \relax 
  \or 
    \ifdim\!ticklocation<-\!!origin
      \edef\!Loc{$\!Loc$}%
    \fi
    \edef\!tickfield{{\the\!ticklocation}{\!Loc}}%
    \expandafter\!listaddon\expandafter{\!tickfield}\!LTlist%
  \or 
    \expandafter\!gettickvaluefrom\!tickvalueslist
    \edef\!tickfield{{\the\!ticklocation}{\!value}}%
    \expandafter\!listaddon\expandafter{\!tickfield}\!LTlist%
  \fi}
\def\!separate#1\!nil{%
  \!ifnextchar{-}{\!!separate}{\!!!separate}#1\!nil}
\def\!!separate-#1\!nil{%
  \def\!sign{-}%
  \!!!!separate#1..\!nil}
\def\!!!separate#1\!nil{%
  \def\!sign{+}%
  \!!!!separate#1..\!nil}
\def\!!!!separate#1.#2.#3\!nil{%
  \def\!arg{#1}%
  \ifx\!arg\!empty
    \!countA=0
  \else
    \!countA=\!arg
  \fi
  \def\!arg{#2}%
  \ifx\!arg\!empty
    \!countB=0
  \else
    \!countB=\!arg
  \fi}
\def\!countfigures#1{%
  \if #1/%
    \def\!next{\ignorespaces}%
  \else
    \multiply\!scalefactor 10
    \def\!next{\!countfigures}%
  \fi
  \!next}
\def\!scaleup#1by#2to#3{%
  \expandafter\!separate#1\!nil
  \multiply\!countA #2\relax
  \advance\!countA \!countB
  \if -\!sign
    \!countA=-\!countA
  \fi
  #3=\!countA
  \ignorespaces}
\def\!scaledown#1by#2to#3{%
  \!countA=#1\relax
  \ifnum \!countA<0 
    \def\!sign{-}
    \!countA=-\!countA
  \else
    \def\!sign{}%
  \fi
  \!countB=\!countA
  \divide\!countB #2\relax
  \!countC=\!countB
    \multiply\!countC #2\relax
  \advance \!countA -\!countC
  \edef#3{\!sign\the\!countB.}
  \!countC=\!countA 
  \ifnum\!countC=0 
    \!countC=1
  \fi
  \multiply\!countC 10
  \!loop \ifnum #2>\!countC
    \edef#3{#3\!zero}%
    \multiply\!countC 10
  \repeat
  \edef#3{#3\the\!countA}
  \ignorespaces}
\def\!placetickvalues{%
  \advance\!offset \tickstovaluesleading
  \if!xswitch
    \setbox\!boxA=\hbox{%
      \def\\##1##2{%
        \!dimenput {##2} [B] (##1,\!axisylevel)}%
      \beginpicture 
        \!LTlist
      \endpicturesave <\!Xsave,\!Ysave>}%
    \!dimenA=\!axisylevel
      \advance\!dimenA -\!Ysave
      \advance\!dimenA \!tickysign\!offset
      \if -\!tickysign
        \advance\!dimenA -\ht\!boxA
      \else
        \advance\!dimenA  \dp\!boxA
      \fi
    \advance\!offset \ht\!boxA 
      \advance\!offset \dp\!boxA
    \!dimenput {\box\!boxA} [Bl] <\!Xsave,\!Ysave> (\!zpt,\!dimenA)
  \else
    \setbox\!boxA=\hbox{%
      \def\\##1##2{%
        \!dimenput {##2} [r] (\!axisxlevel,##1)}%
      \beginpicture 
        \!LTlist
      \endpicturesave <\!Xsave,\!Ysave>}%
    \!dimenA=\!axisxlevel
      \advance\!dimenA -\!Xsave
      \advance\!dimenA \!tickxsign\!offset
      \if -\!tickxsign
        \advance\!dimenA -\wd\!boxA
      \fi
    \advance\!offset \wd\!boxA
    \!dimenput {\box\!boxA} [Bl] <\!Xsave,\!Ysave> (\!dimenA,\!zpt)
  \fi}
\normalgraphs
\catcode`!=12 
\catcode`@=11 \catcode`!=11
\let\!pictexendpicture=\endpicture 
\let\!pictexframe=\frame
\let\!pictexlinethickness=\linethickness
\let\!pictexmultiput=\multiput
\let\!pictexput=\put
\def\beginpicture{%
  \setbox\!picbox=\hbox\bgroup%
  \let\endpicture=\!pictexendpicture
  \let\frame=\!pictexframe
  \let\linethickness=\!pictexlinethickness
  \let\multiput=\!pictexmultiput
  \let\put=\!pictexput
  \let\input=\@@input   
  \!xleft=\maxdimen  
  \!xright=-\maxdimen
  \!ybot=\maxdimen
  \!ytop=-\maxdimen}
\let\frame=\!latexframe
\let\pictexframe=\!pictexframe
\let\linethickness=\!latexlinethickness
\let\pictexlinethickness=\!pictexlinethickness
\let\\=\@normalcr
\catcode`@=12 \catcode`!=12

\makeatother

\def\PP{P\!P}
\def\LP{L\!P}
\def\N{\mbox{I\kern-.2emN}}
\def\lb{\discretionary{}{}{}}
 
\begin{document}

\maketitle

\begin{abstract}

This paper investigates model merging, a technique for deriving Markov
models from text or speech corpora. Models are derived by starting with
a large and specific model and by successively combining states to build
smaller and more general models. We present methods to reduce the time
complexity of the algorithm and report on experiments on deriving
language models for a speech recognition task. The experiments show 
the advantage of model merging over the standard bigram
approach. The merged model assigns a lower perplexity to the test set
and uses considerably fewer states. 

\end{abstract}

\section{Introduction}

    Hidden Markov Models are commonly used for statistical language
models, e.g.\ in part-of-speech tagging and speech recognition
\cite{Rabiner89}. The models need a large set of parameters which are
induced from a (text-) corpus. The parameters should be optimal in the
sense that the resulting models assign high probabilities to seen
training data as well as new data that arises in an application. 

    There are several methods to estimate model parameters. The first
one is to use each word (type) as a state and estimate the transition
probabilities between two or three words by using the relative
frequencies of a corpus. This method is commonly used in speech
recognition and known as word-bigram or word-trigram model. The relative
frequencies have to be smoothed to handle the sparse data problem and to
avoid zero probabilities. 

    The second method is a variation of the first method. Words are
automatically grouped, e.g. by similarity of distribution in the corpus
\cite{Pereira93}. The relative frequencies of pairs or triples of groups
(categories, clusters) are used as model parameters, each group is
represented by a state in the model. The second method has the advantage
of drastically reducing the number of model parameters and thereby
reducing the sparse data problem; there is more data per group than per
word, thus estimates are more precise.

    The third method uses manually defined categories. They are
linguistically motivated and usually called {\em parts-of-speech\/}. An
important difference to the second method with automatically derived
categories is that with the manual definition a word can belong to more
than one category. A corpus is (manually) tagged with the categories and
transition probabilities between two or three categories are estimated
from their relative frequencies. This method is commonly used for
part-of-speech tagging \cite{Church88}.

    The fourth method is a variation of the third method and is also
used for part-of-speech tagging. This method does not need a
pre-annotated corpus for parameter estimation. Instead it uses a lexicon
stating the possible parts-of-speech for each word, a raw text corpus,
and an initial bias for the transition and output probabilities. The
parameters are estimated by using the Baum-Welch algorithm
\cite{Baum70}. The accuracy of the derived model depends heavily on the
initial bias, but with a good choice results are comparable to those of
method three \cite{Cutting92}.

    This paper investigates a fifth method for estimating natural
language models, combining the advantages of the methods mentioned
above. It is suitable for both speech recognition and part-of-speech
tagging, has the advantage of automatically deriving word categories
from a corpus and is capable of recognizing the fact that a word belongs
to more than one category. Unlike other techniques it not only induces
transition and output probabilities, but also the model topology, i.e.,
the number of states, and for each state the outputs that have a
non-zero probability. The method is called model merging and was
introduced by \cite{Omohundro92}.

The rest of the paper is structured as follows. We first give a short
introduction to Markov models and  present the model merging technique.
Then, techniques for reducing the time complexity are presented and we
report two experiments using these techniques.

\section{Markov Models}
\label{secMarkovModels}

    A discrete output, first order {\em Markov Model}
consists of
 \begin{itemize}
 \item	a finite set of states ${\cal Q} \cup \{q_s, q_e\}$, $q_s,q_e
	\not\in {\cal Q}$, with $q_s$
	the start state, and $q_e$ the end state;
 \item	a finite output alphabet $\Sigma$;
 \item	a $(|{\cal Q}|+1)\times(|{\cal Q}|+1)$ matrix, specifying the
	probabilities of state transitions $p(q'|q)$ between states $q$
	and $q'$ (there are no transitions into $q_s$, and no
	transitions originating in $q_e$);
	for each state $q \in {\cal Q}\cup\{q_s\}$, the
	sum of the outgoing transition probabilities is 1, 
	$\sum\limits_{q'\in{\cal Q}\cup\{q_e\}}p(q'|q)=1$;
 \item	a $|{\cal Q}|\times|\Sigma|$ matrix, specifying the output
	probabilities $p(\sigma|q)$ of state $q$ emitting output $\sigma$;
	for each state $q\in{\cal Q}$, the
	sum of the output probabilities is 1,
	$\sum\limits_{\sigma\in\Sigma}p(\sigma|q)=1$.
 \end{itemize}

A Markov model starts running in the start state $q_s$, makes a
transition at each time step, and stops when reaching the end state
$q_e$. The transition from one state to another is done according to the
probabilities specified with the transitions. Each time a state is
entered (except the start and end state) one of the outputs is chosen
(again according to their probabilities) and emitted.

\subsection{Assigning Probabilities to Data}

For the rest of the paper, we are interested in the probabilities
which are assigned to sequences of outputs by the Markov models.
These can be calculated in the following way.

Given a model $M$, a sequence of outputs $\sigma = \sigma_1 \dots
\sigma_k$ and a sequence of states $Q = q_1 \dots q_k$ (of same length),
the probability that the model running through the sequence of states
and emitting the given outputs is
\[
	P_M(Q, \sigma) = \left(\prod_{i=1}^k p_M(q_i|q_{i-1})
		 p_M(\sigma_i|q_i) \right) p_M(q_e|q_i)
\]
(with $q_0 = q_s$).
A sequence of outputs can be emitted by more than one sequence of
states, thus we have to sum over all sequences of states with the given
length to get the probability that a model emits a given sequence of
outputs:
\[
	P_M(\sigma) = \sum_Q P_M(Q,\sigma).
\]
The probabilities are calculated very efficiently with the Viterbi
algorithm \cite{Viterbi67}. Its time complexity is linear to the
sequence length despite the exponential growth of the search space.

\subsection{Perplexity}

Markov models assign rapidly decreasing probabilities to output
sequences of increasing length. To compensate for different lengths and
to make their probabilities comparable, one uses the perplexity $\PP$ of
an output sequence instead of its probability. The perplexity is defined
as
\[
	\PP_M(\sigma) = \frac{1}{\sqrt[k]{P_M(\sigma)}}.
\]
The probability is normalized by taking the $k^{th}$ root ($k$ is the
length of the sequence). Similarly, the log perplexity $\LP$ is
defined:
\[
	\LP_M(\sigma) = \log PP_M(\sigma) = \frac{-\log P_M(\sigma)}{k}.
\]
Here, the log probability is normalized by dividing by the length of the
sequence. 

$\PP$ and $\LP$ are defined such that higher perplexities (log
perplexities, resp.) correspond to lower probabilities, and vice versa.
These measures are used to determine the quality of Markov models. The
lower the perplexity (and log perplexity) of a test sequence, the higher
its probability, and thus the better it is predicted by the model.

\section{Model Merging}
\label{secModelMerging}

    Model merging is a technique for inducing model parameters for
Markov models from a text corpus. It was introduced in 
\cite{Omohundro92} and \cite{Stolcke94a} to induce models for regular
languages from a few samples, and adapted to natural language models in
\cite{Brants95b}. Unlike other techniques it not only induces transition
and output probabilities from the corpus, but also the model topology,
i.e., the number of states and for each state the outputs that have
non-zero probability. In $n$-gram approaches the topology is fixed.
E.g., in a pos-$n$-gram model, the states are mostly syntactically
motivated, each state represents a syntactic category and only words
belonging to the same category have a non-zero output probability in a
particular state. However the $n$-gram-models make the implicit
assumption that all words belonging to the same category have a similar
distribution in a corpus. This is not true in most of the cases. 

\begin{figure*}
\hrule
\begin{center}
\def\kreis(#1,#2)#3{\put(#1,#2){\circle{10}\makebox(0,0){#3}}}
\setlength{\unitlength}{0.75mm}
\begin{picture}(160,60)
\put(2,52){\makebox(0,0)[lb]{\bf a)}}
\normalsize
\kreis(5,25){\scriptsize Start}
\kreis(35,5){5}
\kreis(35,25){3}
\kreis(35,45){1}
\kreis(65,5){6}
\kreis(65,25){4}
\kreis(65,45){2}
\kreis(95,5){7}
\kreis(125,5){8}
\kreis(155,25){\scriptsize End}
\footnotesize
\put(10.5,25){\vector(1,0){19}}
\put(20,24){\makebox(0,0)[t]{0.33}}
\put(10,28.33){\vector(3,2){20}}
\put(20,36){\makebox(0,0)[rb]{0.33}}
\put(10,21.67){\vector(3,-2){20}}
\put(20,14){\makebox(0,0)[rt]{0.33}}
\put(40.5,5){\vector(1,0){19}}
\put(40.5,25){\vector(1,0){19}}
\put(40.5,45){\vector(1,0){19}}
\put(70.5,5){\vector(1,0){19}}
\put(70.5,25){\vector(1,0){79}}
\put(70.5,45){\line(1,0){54.5}}
\put(125,45){\vector(3,-2){25}}%
\put(100.5,5){\vector(1,0){19}}
\put(130,8.33){\vector(3,2){20}}
\small
\put(35,52){\makebox(0,0)[b]{\em a}}
\put(65,52){\makebox(0,0)[b]{\em b}}
\put(35,32){\makebox(0,0)[b]{\em a}}
\put(65,32){\makebox(0,0)[b]{\em c}}
\put(35,12){\makebox(0,0)[b]{\em a}}
\put(65,12){\makebox(0,0)[b]{\em b}}
\put(95,12){\makebox(0,0)[b]{\em a}}
\put(125,12){\makebox(0,0)[b]{\em c}}
\normalsize
\put(162,52){\small\makebox(0,0)[rb]{$p(S|\mbox{M}_a) = \frac{1}{27} 
	\simeq 3.7\cdot 10^{-2}$}}
\end{picture}

\bigskip

\begin{picture}(160,60)
\put(2,52){\makebox(0,0)[lb]{\bf b)}}
\normalsize
\kreis(5,25){\scriptsize Start}
\kreis(35,5){5}
\kreis(35,25){1,3}
\kreis(65,5){6}
\kreis(65,25){4}
\kreis(65,45){2}
\kreis(95,5){7}
\kreis(125,5){8}
\kreis(155,25){\scriptsize End}
\footnotesize
\put(10.5,25){\vector(1,0){19}}
\put(20,24){\makebox(0,0)[t]{0.67}}
\put(10,22.5){\vector(3,-2){20.5}}
\put(20,14.5){\makebox(0,0)[rt]{0.33}}
\put(40.5,5){\vector(1,0){19}}
\put(40.5,25){\vector(1,0){19}}
\put(50,24){\makebox(0,0)[t]{0.5}}

\put(40,28.33){\vector(3,2){20}}
\put(50,36){\makebox(0,0)[rb]{0.5}}

\put(70.5,5){\vector(1,0){19}}
\put(70.5,25){\vector(1,0){79}}
\put(70.5,45){\line(1,0){54.5}}
\put(125,45){\vector(3,-2){25}}%
\put(100.5,5){\vector(1,0){19}}
\put(130,8.33){\vector(3,2){20}}
\small
\put(65,52){\makebox(0,0)[b]{\em b}}
\put(35,32){\makebox(0,0)[b]{\em a}}
\put(65,32){\makebox(0,0)[b]{\em c}}
\put(35,12){\makebox(0,0)[b]{\em a}}
\put(65,12){\makebox(0,0)[b]{\em b}}
\put(95,12){\makebox(0,0)[b]{\em a}}
\put(125,12){\makebox(0,0)[b]{\em c}}
\normalsize
\put(162,52){\small\makebox(0,0)[rb]{$p(S|\mbox{M}_b) = \frac{1}{27}
	\simeq 3.7\cdot 10^{-2}$}}
\end{picture}

\bigskip

\begin{picture}(160,40)(0,-10)
\put(2,22){\makebox(0,0)[lb]{\bf c)}}
\normalsize
\kreis(5,5){\scriptsize Start}
\kreis(35,5){\footnotesize 1,3,5}
\kreis(65,5){2,6}
\kreis(95,5){7}
\kreis(125,5){4,8}
\kreis(155,5){\scriptsize End}
\footnotesize
\put(10.5,5){\vector(1,0){19}}
\put(40.5,5){\vector(1,0){19}}
\put(50,6.5){\makebox(0,0)[b]{0.67}}
\put(39,9){\line(2,1){22}}
\put(61,20){\line(1,0){38}}
\put(99,20){\vector(2,-1){22}}
\put(50,16){\makebox(0,0)[rb]{0.33}}
\put(70.5,5){\vector(1,0){19}}
\put(80,6.5){\makebox(0,0)[b]{0.5}}
\put(69,1){\line(2,-1){22}}
\put(91,-10){\line(1,0){38}}
\put(129,-10){\vector(2,1){22}}
\put(110,-9){\makebox(0,0)[b]{0.5}}
\put(100.5,5){\vector(1,0){19}}
\put(130.5,5){\vector(1,0){19}}
\small
\put(35,12){\makebox(0,0)[b]{\em a}}
\put(65,12){\makebox(0,0)[b]{\em b}}
\put(95,12){\makebox(0,0)[b]{\em a}}
\put(125,12){\makebox(0,0)[b]{\em c}}
\normalsize
\put(162,22){\small\makebox(0,0)[rb]{$p(S|\mbox{M}_c) = \frac{1}{27}
	\simeq 3.7\cdot 10^{-2}$}}
\end{picture}

\bigskip

\begin{picture}(160,40)(0,-10)
\put(2,22){\makebox(0,0)[lb]{\bf d)}}
\normalsize
\kreis(5,5){\scriptsize Start}
\kreis(35,5){\scriptsize\shortstack{1,3\\[-.8ex]5,7}}
\kreis(65,5){2,6}
\kreis(125,5){4,8}
\kreis(155,5){\scriptsize End}
\footnotesize
\put(10.5,5){\vector(1,0){19}}
\put(40.5,7){\vector(1,0){19}}
\put(50,8){\makebox(0,0)[b]{0.5}}
\put(39,9){\line(2,1){22}}
\put(61,20){\line(1,0){38}}
\put(99,20){\vector(2,-1){22}}
\put(50,16){\makebox(0,0)[rb]{0.5}}
\put(59.5,3){\vector(-1,0){19}}
\put(50,2){\makebox(0,0)[t]{0.5}}
\put(69,1){\line(2,-1){22}}
\put(91,-10){\line(1,0){38}}
\put(129,-10){\vector(2,1){22}}
\put(110,-9){\makebox(0,0)[b]{0.5}}
\put(130.5,5){\vector(1,0){19}}
\small
\put(35,12){\makebox(0,0)[b]{\em a}}
\put(65,12){\makebox(0,0)[b]{\em b}}
\put(125,12){\makebox(0,0)[b]{\em c}}
\normalsize
\put(162,22){\small\makebox(0,0)[rb]{$p(S|\mbox{M}_d) = \frac{1}{64}
	\simeq 1.6\cdot 10^{-2}$}}
\end{picture}

\bigskip

\begin{picture}(160,30)
\put(2,22){\makebox(0,0)[lb]{\bf e)}}
\normalsize
\kreis(5,5){\scriptsize Start}
\kreis(35,5){\scriptsize\shortstack{1,3\\[-.8ex]5,7}}
\kreis(65,5){\scriptsize\shortstack{2,4\\[-.8ex]6,8}}
\kreis(155,5){\scriptsize End}
\footnotesize
\put(10.5,5){\vector(1,0){19}}
\put(40,7){\vector(1,0){20}}
\put(60,3){\vector(-1,0){20}}
\put(50,2){\makebox(0,0)[t]{0.25}}
\put(70.5,5){\vector(1,0){79}}
\put(110,6){\makebox(0,0)[b]{0.75}}
\small
\put(35,12){\makebox(0,0)[b]{\em a}}
\put(61.5,9.5){\vector(-1,1){6}}
\put(55,16){\makebox(0,0)[b]{\small\em b}}
\put(58.5,12.5){\makebox(0,0)[lb]{\scriptsize\em 0.5}}
\put(68.5,9.5){\vector(1,1){6}}
\put(75,16){\makebox(0,0)[b]{\small\em c}}
\put(71.5,12.5){\makebox(0,0)[rb]{\scriptsize\em 0.5}}
\normalsize
\put(162,22){\small\makebox(0,0)[rb]{$p(S|\mbox{M}_e) = \frac{27}{4096}
	\simeq 6.6\cdot 10^{-3}$}}
\end{picture}
\end{center}

\hrule
\caption{Model merging for a corpus $S = \{ab, ac, abac\}$, starting with
	the trivial model in a) and ending with the generalization
	$(a(b|c))^+$ in e). Several steps of merging between model b)
	and c) are not shown. Unmarked transitions and outputs have
	probability 1.}
\label{figModelMerging}
\end{figure*}
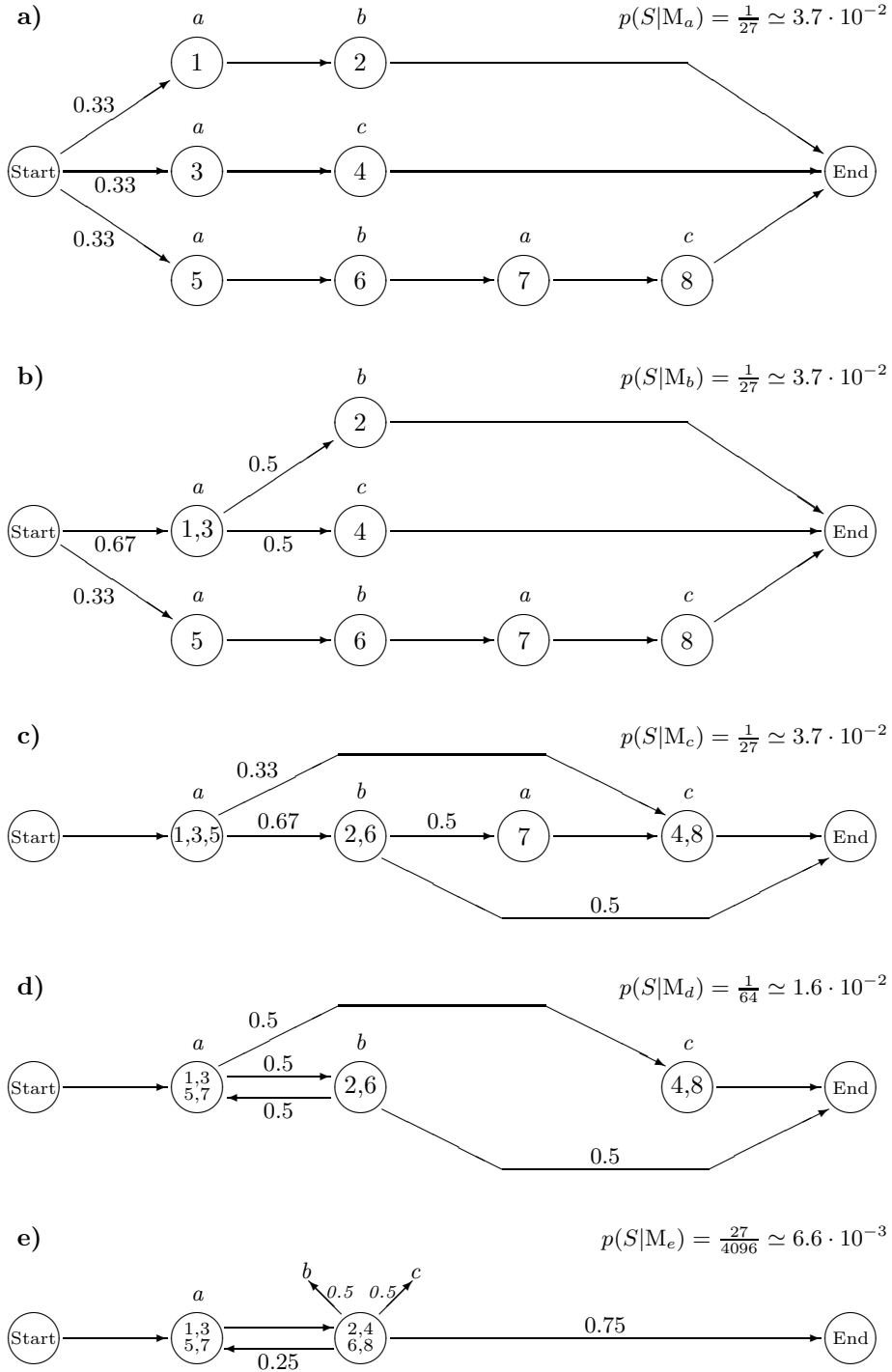

    By estimating the topology, model merging groups words into
categories, since all words that can be emitted by the same state form a
category. The advantage of model merging in this respect is that it can
recognize that a word (the type) belongs to more than one category, while
each occurrence (the token) is assigned a unique category. This
naturally reflects manual syntactic categorizations, where a word can
belong to several syntactic classes but each occurrence of a word is
unambiguous.

\subsection{The Algorithm}

    Model merging induces Markov models in the following way. Merging
starts with an initial, very general model. For this purpose, the
maximum likelihood Markov model is chosen, i.e., a model that exactly
matches the corpus. There is one path for each utterance in the corpus
and each path is used by one utterance only. Each path gets the same
probability $1/u$, with $u$ the number of utterances in the corpus. This
model is also referred to as the trivial model. Figure
\ref{figModelMerging}.a shows the trivial model for a corpus with words
$a, b, c$ and utterances $ab, ac, abac$. It has one path for each of the
three utterances $ab$, $ac$, and $abac$, and each path gets the same
probability $1/3$. The trivial model assigns a probability of
$p(S|\mbox{M}_a) = 1/27$ to the corpus. Since the model makes an
implicit independence assumption between the utterances, the corpus
probability is calculated by multiplying the utterance's probabilities,
yielding 
$1/3 \cdot 1/3 \cdot 1/3 = 1/27$.

    Now states are merged successively, except for the start and end
state. Two states are selected and removed and a new merged state is
added. The transitions from and to the old states are redirected to the
new state, the transition probabilities are adjusted to maximize the
likelihood of the corpus; the outputs are joined and their probabilities
are also adjusted to maximize the likelihood. One step of merging can be
seen in figure \ref{figModelMerging}.b. States 1 and 3 are removed, a
combined state 1,3 is added, and the probabilities are adjusted.

    The criterion for selecting states to merge is the probability of
the Markov model generating the corpus. We want this probability to stay
as high as possible. Of all possible merges (generally, there are
$k(k-1)/2$ possible merges, with $k$ the number of states exclusive
start and end state which are not allowed to merge) we take the merge
that results in the minimal change of the probability. For the trivial
model and $u$ pairwise different utterances the probability is
$p(S|M_{triv}) = 1/u^u$. The probability either stays constant, as in
Figure \ref{figModelMerging}.b and c, or decreases, as in
\ref{figModelMerging}.d and e. The probability never increases
because the trivial model is the maximum likelihood model, i.e., it
maximizes the probability of the corpus given the model.

    Model merging stops when a predefined threshold for the corpus
probability is reached. Some statistically motivated criteria for
termination using model priors are discussed in \cite{Stolcke94a}.

\section{Using Model Merging}

    The model merging algorithm needs several optimizations to be
applicable to large natural language corpora, otherwise the amount of
time needed for deriving the models is too large. Generally, there are
$O(l^2)$ hypothetical merges to be tested for each merging step ($l$ is
the length of the training corpus). The probability of the training
corpus has to be calculated for each hypothetical merge, which is $O(l)$
with dynamic programming. Thus, each step of merging is $O(l^3)$. If we
want to reduce the model from size $l+2$ (the trivial model, which
consists of one state for each token plus initial and final states) to
some fixed size, we need $O(l)$ steps of merging. Therefore, deriving a
Markov model by model merging is $O(l^4)$ in time.

\cite{Stolcke94a} discuss several computational shortcuts and
approximations:
\begin{enumerate}
\item	Immediate merging of identical initial and final states of
	different utterances. These merges do not change the corpus
	probability and thus are the first merges anyway.
\item	Usage of the Viterbi path (best path) only instead of summing up
	all paths to determine the corpus probability.
\item	The assumption that all
	input samples retain their Viterbi path after merging.
	Making this approximation, it is no
	longer necessary to re-parse the whole corpus for each
	hypothetical merge.
\end{enumerate}

We use two additional strategies to reduce the time complexity of the
algorithm: a series of cascaded constraints on the merges and the
variation of the starting point.

\subsection{Constraints}

    When applying model merging one can observe that first mainly states
with the same output are merged. After several steps of merging, it is
no longer the same output but still mainly states that output words of
the same syntactic category are merged. This behavior can be exploited
by introducing constraints on the merging process. The constraints allow
only some of the otherwise possible merges. Only the allowed merges are
tested for each step of merging.

We consider constraints that divide the states of the current model into
equivalence classes. Only states belonging to the same class are allowed
to merge. E.g., we can divide the states into classes generating the
same outputs. If the current model has $N$ states and we divide them
into $k>1$ nonempty equivalence classes $C_1\dots C_k$, then, instead of
$N(N-1)/2$, we have to test
\[
	\sum_{i=1}^k \frac{|C_i|(|C_i|-1)}{2} < \frac{N(N-1)}{2}
\]
merges only.

The best case for a model of size $N$ is the division into $N/2$
classes of size $2$. Then, only $N/2$ merges must be tested to find the
best merge.

The best division into $k > 1$ classes for some model of size $N$ is the
creation of classes that all have the same size $N/k$ (or an
approximation if $N/k \not\in \N$). Then, 
\[
	\frac{\frac{N}{k}(\frac{N}{k}-1)}{2}\cdot k =
	\frac{N(\frac{N}{k}-1)}{2}
\]
must be tested for each step of merging.

Thus, the introduction of these constraints does not reduce the order of
the time complexity, but it can reduce the constant factor significantly
(see section about experiments). 

The following equivalence classes can be used for constraints 
when using untagged corpora:
\begin{enumerate}
\item	States that generate the same outputs (unigram constraint)
\item	unigram constraint, and additionally all predecessor states must
	generate the same outputs (bigram constraint)
\item	trigrams or higher, if the corpora are large enough
\item	a variation of one: states that output words belonging to one
	ambiguity class, i.e. can be of a certain number of syntactic
	classes. 
\end{enumerate}	

Merging starts with one of the constraints. After a number of merges
have been performed, the constraint is discarded and a weaker one is
used instead.

The standard $n$-gram approaches are special cases of using model
merging and constraints. E.g., if we use the unigram constraint, and
merge states until no further merge is possible under this constraint,
the resulting model is a standard bigram model, regardless of the order
in which the merges were performed.

In practice, a constraint will be discarded before no further merge is
possible (otherwise the model could have been derived directly, e.g., by
the standard $n$-gram technique). Yet, the question when to discard a
constraint to achieve best results is unsolved.

\subsection{The Starting Point}

The initial model of the original model merging procedure is the maximum
likelihood or trivial model. This model has the advantage
of directly representing the corpus. But its disadvantage is its huge
number of states. A lot of computation time can be saved by choosing an
initial model with fewer states. 

The initial model must have two properties:
\begin{enumerate}
\item	it must be larger than the intended model, and
\item	it must be easy to construct.
\end{enumerate}
The trivial model has both properties. A class of models that
can serve as the initial model as well are $n$-gram models.
These models are smaller by one or more orders of magnitude than the
trivial model and therefore could speed up the derivation of a model
significantly. 

This choice of a starting point excludes a lot of solutions which are
allowed when starting with the maximum likelihood model. Therefore,
starting with an $n$-gram model yields a model that is at most
equivalent to one that is generated when starting with the trivial
model, and that can be much worse. But it should be still better than
any $n$-gram model that is of lower of equal order than the initial
model.

\section{Experiments}
\label{secExperiment}

\subsection{Model Merging vs.\ Bigrams}

The first experiment compares model merging with a standard bigram
model. Both are trained on the same data. We use $N_{train} = 14,421$
words of the Verbmobil corpus. The corpus consists of transliterated
dialogues on business appointments\footnote{Many thanks to the Verbmobil
project for providing these data. We use dialogues that were recorded in
1993 and 94, and which are now available from the Bavarian Archive for
Speech Sig\-nals BAS (http://\lb www.phonetik.uni-muenchen.de/\lb
Bas/\lb BasHomeeng.html).}. The models are tested on $N_{test}=2,436$
words of the same corpus. Training and test parts are disjunct.

\begin{figure*}
\hrule
\bigskip

\hspace*{0pt}
\beginpicture
\setcoordinatesystem units <.7mm,1.7mm>
\setplotarea x from 0 to 146, y from 0 to 36.5
\axis bottom ticks 
	withvalues 0 1 2 3 4 5 6 7 8 9 10 11 12 13 14 /
	from 0 to 140 by 10 /
\axis bottom ticks short from 5 to 135 by 10 /
\put {$\times 10^3$ merges} <0mm,-2.5mm> [lt] at 147 0

\axis bottom shiftedto y=-5 ticks
	withvalues 14 13 12 11 10 9 8 7 6 5 4 3 2 1 0 /
	at 4.23 14.23 24.23 34.23 44.23 54.23 64.23 74.23 84.23 94.23 104.23 114.23 124.23 134.23 144.23 / /
\put {$\times 10^3$ states} [lt] <0mm,-2.5mm> at 147 -5

\axis left ticks
	withvalues 0.5 1.0 1.5 2.0 2.5 /
	at 7.21 14.42 21.63 28.84 36.05 / /
\axis left ticks short
	at 1.44  2.88  4.33  5.77  8.65 10.09 11.54 12.98 15.86 17.31
	   18.75 20.19 23.07 24.52 25.96 27.40 30.28 31.73 33.17 34.61 / /
\put {$-\log_{10} p / N_{\mbox{\scriptsize train}}$} [b] at 0 37.5
\linethickness=1.5mm
\setlinear
\setplotsymbol({\rule{.8pt}{.8pt}})
\plot
	0	 1.7075
	14	 1.7075
	15	 1.7389
	16	 1.8163
	17	 1.8883
	18	 1.9534
	19	 2.0098
	20	 2.0622
	21	 2.1147
	22	 2.1814
	23	 2.2401
	24	 2.3045
	25	 2.3601
	26	 2.4165
	27	 2.4706
	28	 2.5231
	29	 2.5786
	30	 2.6322
	31	 2.6856
	32	 2.7403
	33	 2.7961
	34	 2.8531
	35	 2.8996
	36	 2.9556
	37	 3.0081
	38	 3.0639
	39	 3.1189
	40	 3.1689
	41	 3.2170
	42	 3.2718
	43	 3.3251
	44	 3.3775
	45	 3.4268
	46	 3.4753
	47	 3.5253
	48	 3.5741
	49	 3.6212
	50	 3.6707
	51	 3.7136
	52	 3.7631
	53	 3.8168
	54	 3.8678
	55	 3.9120
	56	 3.9564
	57	 4.0072
	58	 4.0670
	59	 4.1202
	60	 4.1728
	61	 4.2239
	62	 4.2731
	63	 4.3269
	64	 4.3755
	65	 4.4310
	66	 4.4775
	67	 4.5192
	68	 4.5732
	69	 4.6239
	70	 4.6732
	71	 4.7258
	72	 4.7776
	73	 4.8302
	74	 4.8759
	75	 4.9286
	76	 4.9816
	77	 5.0323
	78	 5.0844
	79	 5.1377
	80	 5.2177
	81	 5.2964
	82	 5.3694
	83	 5.4405
	84	 5.5196
	85	 5.6293
	86	 5.7378
	87	 5.8485
	88	 5.9591
	89	 6.0658
	90	 6.1649
	91	 6.2729
	92	 6.3862
	93	 6.5007
	94	 6.6142
	95	 6.7282
	96	 6.8409
	97	 6.9567
	98	 7.0644
	99	 7.1778
	100	 7.2852
	101	 7.4011
	102	 7.5127
	103	 7.6289
	104	 7.7477
	105	 7.8918
	106	 8.0256
	107	 8.1826
	108	 8.3440
	109	 8.5169
	110	 8.6964
	111	 8.8957
	112	 9.0807
	113	 9.2833
	114	 9.5090
	115	 9.7344
	116	 9.9711
	117	10.2040
	118	10.4439
	119	10.6891
	120	10.9831
	121	11.2966
	122	11.6608
	123	12.0597
	124	12.4951
	125	12.9675
	126	13.0218
	127	13.0851
	128	13.1711
	129	13.2827
	130	13.4380
	131	13.6557
	132	13.9411
	133	14.3525
	134	14.8646
	135	15.4545
	136	16.1519
	137	16.9282
	138	17.7940
	139	18.7785
	140	19.9588
	141	21.3818
	142	23.4625
	143	27.2294
	143.87	34.7349
/
\setplotsymbol({\rule{.1pt}{.1pt}})
\plot
	125	12.9675
	126	13.4973
	127	14.1913
	128	15.0103
	129	16.1523
	129.83	17.3589
/
\put {\boldmath$lp$} [lt] at 145 34.7349

\setplotsymbol({\rule{.4pt}{.4pt}})
\arrow <7pt> [.2,.67] from 110 14.5 to 122 13.3
\put {\shortstack[r]{constraint\\change}} [rb] at 110 14.5

\setplotsymbol({\rule{.4pt}{.4pt}})
\plot
	14	0.0000
	15	0.3140
	16	0.7740
	17	0.7200
	18	0.6510
	19	0.5640
	20	0.5240
	21	0.5250
	22	0.6670
	23	0.5870
	24	0.6440
	25	0.5560
	26	0.5640
	27	0.5410
	28	0.5250
	29	0.5550
	30	0.5360
	31	0.5340
	32	0.5470
	33	0.5580
	34	0.5700
	35	0.4650
	36	0.5600
	37	0.5250
	38	0.5580
	39	0.5500
	40	0.5000
	41	0.4810
	42	0.5480
	43	0.5330
	44	0.5240
	45	0.4930
	46	0.4850
	47	0.5000
	48	0.4880
	49	0.4710
	50	0.4950
	51	0.4290
	52	0.4950
	53	0.5370
	54	0.5100
	55	0.4420
	56	0.4440
	57	0.5080
	58	0.5980
	59	0.5320
	60	0.5260
	61	0.5110
	62	0.4920
	63	0.5380
	64	0.4860
	65	0.5550
	66	0.4650
	67	0.4170
	68	0.5400
	69	0.5070
	70	0.4930
	71	0.5260
	72	0.5180
	73	0.5260
	74	0.4570
	75	0.5270
	76	0.5300
	77	0.5070
	78	0.5210
	79	0.5330
	80	0.8000
	81	0.7870
	82	0.7300
	83	0.7110
	84	0.7910
	85	1.0970
	86	1.0850
	87	1.1070
	88	1.1060
	89	1.0670
	90	0.9910
	91	1.0800
	92	1.1330
	93	1.1450
	94	1.1350
	95	1.1400
	96	1.1270
	97	1.1580
	98	1.0770
	99	1.1340
	100	1.0740
	101	1.1590
	102	1.1160
	103	1.1620
	104	1.1880
	105	1.4410
	106	1.3380
	107	1.5700
	108	1.6140
	109	1.7290
	110	1.7950
	111	1.9930
	112	1.8500
	113	2.0260
	114	2.2570
	115	2.2540
	116	2.3670
	117	2.3290
	118	2.3990
	119	2.4520
	120	2.9400
	121	3.1350
	122	3.6420
	123	3.9890
	124	4.3540
	125	4.7240
	125	0.5430
	126	0.5430
	127	0.6330
	128	0.8600
	129	1.1160
	130	1.5530
	131	2.1770
	132	2.8540
	133	4.1140
	134	5.1210
	135	5.8990
	136	6.9740
	137	7.7630
	138	8.6580
	139	9.8450
	140	11.8030
	141	14.2300
	142	20.8070
/
\setplotsymbol({\rule{.1pt}{.1pt}})
\plot
	125	4.7240
	126	5.2980
	127	6.9400
	128	8.1900
	129	11.4200
/
\put {$dlp$} [lt] at 145 20.8070

\endpicture

\bigskip

\hrule
\setbox0=\hbox{$N_{\mbox{\scriptsize train}}=14,421$}
\caption{Log Perplexity of the training part during merging. Constraints:
	same output until 12,500 / none after 12,500. The thin lines
	show the further development if we retain the the same-output
	constraint until no further merge is possible. The length of the
training part is \copy0.}
\label{figTrainProb}

\bigskip
\bigskip
\bigskip

\hrule
\bigskip

\hspace*{0pt}
\beginpicture
\setcoordinatesystem units <.7mm,60mm>
\setplotarea x from 0 to 145, y from 2.15 to 2.83
\axis bottom ticks 
	withvalues 0 1 2 3 4 5 6 7 8 9 10 11 12 13 14 /
	from 0 to 140 by 10 /
\axis bottom ticks short from 5 to 135 by 10 /
\put {$\times 10^3$ merges} [lt] <0mm,-2.5mm> at 147 2.15

\axis bottom shiftedto y=2 ticks
	withvalues 14 13 12 11 10 9 8 7 6 5 4 3 2 1 0 /
	at 4.23 14.23 24.23 34.23 44.23 54.23 64.23 74.23 84.23 94.23
	   104.23 114.23 124.23 134.23 144.23 / /
\put {$\times 10^3$ states} [lt] <0mm,-2.5mm> at 147 2

\axis left ticks 
	numbered from 2.2 to 2.8 by .1 /
\put {$-\log_{10} p/N_{\mbox{\scriptsize test}}$} [b] at 0 2.85

\linethickness=1.5mm
\setlinear
\setplotsymbol({\rule{.8pt}{.8pt}})
\plot
	  0.00	 2.7893
	  1.00	 2.7893
	  2.00	 2.7893
	  3.00	 2.7893
	  4.00	 2.7893
	  5.00	 2.7893
	  6.00	 2.7893
	  7.00	 2.7893
	  8.00	 2.7893
	  9.00	 2.7893
	 10.00	 2.7893
	 11.00	 2.7893
	 12.00	 2.7893
	 13.00	 2.7893
	 14.00	 2.7893
	 15.00	 2.7895
	 16.00	 2.7883
	 17.00	 2.7883
	 18.00	 2.7885
	 19.00	 2.7889
	 20.00	 2.7886
	 20.00	 2.7886
	 21.00	 2.7882
	 22.00	 2.7906
	 23.00	 2.7893
	 24.00	 2.7907
	 25.00	 2.7900
	 26.00	 2.7898
	 27.00	 2.7899
	 28.00	 2.7900
	 29.00	 2.7900
	 30.00	 2.7894
	 31.00	 2.7889
	 32.00	 2.7892
	 33.00	 2.7872
	 34.00	 2.7871
	 35.00	 2.7871
	 36.00	 2.7874
	 37.00	 2.7863
	 38.00	 2.7863
	 39.00	 2.7846
	 40.00	 2.7849
	 41.00	 2.7764
	 42.00	 2.7763
	 43.00	 2.7765
	 44.00	 2.7768
	 45.00	 2.7776
	 46.00	 2.7775
	 47.00	 2.7778
	 48.00	 2.7742
	 49.00	 2.7748
	 50.00	 2.7745
	 51.00	 2.7745
	 52.00	 2.7694
	 53.00	 2.7698
	 54.00	 2.7695
	 55.00	 2.7679
	 56.00	 2.7680
	 57.00	 2.7680
	 58.00	 2.7692
	 59.00	 2.7690
	 60.00	 2.7679
	 60.00	 2.7679
	 61.00	 2.7673
	 62.00	 2.7676
	 63.00	 2.7677
	 64.00	 2.7678
	 65.00	 2.7676
	 66.00	 2.7678
	 67.00	 2.7677
	 68.00	 2.7684
	 69.00	 2.7658
	 70.00	 2.7648
	 71.00	 2.7649
	 72.00	 2.7637
	 73.00	 2.7645
	 74.00	 2.7650
	 75.00	 2.7656
	 76.00	 2.7659
	 77.00	 2.7667
	 78.00	 2.7658
	 79.00	 2.7636
	 80.00	 2.7648
	 81.00	 2.7642
	 82.00	 2.7645
	 83.00	 2.7654
	 84.00	 2.7627
	 85.00	 2.7612
	 86.00	 2.7599
	 87.00	 2.7600
	 88.00	 2.7615
	 89.00	 2.7563
	 90.00	 2.7558
	 91.00	 2.7551
	 92.00	 2.7557
	 93.00	 2.7559
	 94.00	 2.7561
	 95.00	 2.7562
	 96.00	 2.7572
	 97.00	 2.7583
	 98.00	 2.7565
	 99.00	 2.7560
	100.00	 2.7558
	101.00	 2.7542
	102.00	 2.7527
	103.00	 2.7534
	104.00	 2.7506
	105.00	 2.7514
	106.00	 2.7503
	107.00	 2.7469
	108.00	 2.7475
	109.00	 2.7429
	110.00	 2.7431
	111.00	 2.7414
	112.00	 2.7365
	113.00	 2.7255
	114.00	 2.7219
	115.00	 2.7218
	116.00	 2.7206
	117.00	 2.7171
	118.00	 2.7030
	119.00	 2.6964
	120.00	 2.6898
	121.00	 2.6853
	122.00	 2.6598
	123.00	 2.6532
	124.00	 2.6451
	125.00	 2.6072
	126.00	 2.6074
	127.00	 2.6082
	128.00	 2.6087
	129.00	 2.6082
	130.00	 2.6070
	131.00	 2.6086
	132.00	 2.6107
	133.00	 2.6069
	134.00	 2.5994
	135.00	 2.5695
	136.00	 2.5438
	137.00	 2.5283
	138.00	 2.5119
	139.00	 2.4530
	140.00	 2.4129
	141.00	 2.3908
	142.00	 2.3431
	143.00	 2.2587
	143.10	 2.2548
	143.20	 2.2595
	143.30	 2.2620
	143.40	 2.2684
	143.50	 2.2712
	143.60	 2.2979
	143.70	 2.3196
	143.80	 2.3517
	143.87	 2.4679
/
\put {\boldmath$lp$} [lb] <0pt,1mm> at 138 2.6

\setplotsymbol({\rule{.1pt}{.1pt}})
\plot
	125.00	 2.6072
	126.00	 2.5838
	127.00	 2.5599
	128.00	 2.5270
	129.00	 2.4858
	129.83	 2.4030
/

\linethickness=.4pt

\putrule from 0 2.4030 to 145 2.4030
\put {$lp_{\mbox{\scriptsize bigram}}$ (1440 states)} [l] at 147 2.4030
\put {$\circ$} at 129.83 2.4030

\putrule from 0 2.2548 to 145 2.2548
\put {$lp_{\mbox{\scriptsize min}}$ (113 states)} [l] at 147 2.2548
\put {$\circ$} at 143.10 2.2548

\setplotsymbol({\rule{.4pt}{.4pt}})
\arrow <7pt> [.2,.67] from 135 2.75 to 128 2.65
\put {\shortstack[l]{constraint\\change}} [lb] at 135 2.75

\endpicture

\bigskip

\hrule
\setbox0=\hbox{$N_{\mbox{\scriptsize test}}=2,436$}
\caption{Log Perplexity of Test Part During Merging. Constraints:
Same Output until 12,500 / none after 12,500. The thin line shows the
further development if we retain the same-output constraint, finally
yielding a bigram model. The length of the test part is \copy0.}
\label{figTestProb}
\end{figure*}

The bigram model yields a Markov model with 1,440 states. It assigns a
log perplexity of 1.20 to the training part and 2.40 to the
test part.

Model merging starts with the maximum likelihood model for the training
part. It has 14,423 states, which correspond to the 14,421 words (plus
an initial and a final state). The initial log perplexity of the
training part is 0.12. This low value shows that the initial model is
very specialized in the training part. 

We start merging with the same-output (unigram) constraint to reduce
computation time.  After 12,500 merges the constraint is discarded and
from then on all remaining states are allowed to merge. The constraints
and the point of changing the constraint are chosen for pragmatic
reasons. We want the constraints to be as week as possible to allow the
maximal number of solutions but at the same time the number of merges
must be manageable by the system used for computation (a SparcServer1000
with 250MB main memory). As the following experiment will show, the
exact points of introducing/discarding constraints is not important for
the resulting model. 

There are $N_{train}(N_{train}-1)/2\sim 10^8$ hypothetical first merges
in the unconstraint case. This number is reduced to $\sim 7\cdot 10^5$
when using the unigram constraint, thus by a factor of $\sim 150$. By
using the constraint we need about a week of computation time on a
SparcServer 1000 for the whole merging process. Computation would not
have been feasible without this reduction.

Figure \ref{figTrainProb} shows the increase in perplexity during
merging. There is no change during the first 1,454 merges. Here, only
identical sequences of initial and final states are merged (compare
figure \ref{figModelMerging}.a to c). These merges do not influence the
probability assigned to the training part and thus do not change the
perplexity.

Then, perplexity slowly increases. It can never decrease: the maximum
likelihood model assigns the highest probability to the training part
and thus the lowest perplexity. 

Figure \ref{figTrainProb} also shows the perplexity's slope. It is low
until about 12,000 merges, then drastically increases. At about this
point, after 12,500 merges, we discard the constraint. For this reason,
the curve is discontinuous at 12,500 merges. The effect of further
retaining the constraint is shown by the thin lines. These stop after
12,983 merges, when all states with the same outputs are merged (i.e.,
when a bigram model is reached). Merging with\-out a constraint continues
until only three states remain: the initial and the final state plus one
proper state.

Note that the perplexity changes very slowly for the largest part, and
then changes drastically during the last merges. There is a constant
phase between 0 and 1,454 merges. Between 1,454 and $\sim$11,000 merges
the log perplexity roughly linearly increases with the number of merges,
and it explodes afterwards.

What happens to the test part? Model merging starts with a very special
model which then is generalized. Therefore, the perplexity of some
random sample of dialogue data (what the test part is supposed to be)
should decrease during merging. This is exactly what we find in the
experiment. 

Figure \ref{figTestProb} shows the log perplexity of the test part during
merging. Again, we find the discontinuity at the point where the
constraint is changed. And again, we find very little change in
perplexity during about 12,000 initial merges, and large changes during
the last merges.

Model merging finds a model with 113 states, which assigns a log
perplexity of 2.26 to the test part. Thus, in addition to finding a
model with lower log perplexity than the bigram model (2.26 vs. 2.40),
we find a model that at the same time has less than 1/10 of the states
(113 vs.\ 1,440).

To test if we found a model that predicts new data better than the
bigram model and to be sure that we did not find a model that is simply
very specialized to the test part, we use a new, previously unseen part
of the Verbmobil corpus. This part consists of 9,784 words. The bigram
model assigns a log perplexity of 2.78, the merged model with 113 states
assigns a log perplexity of 2.41 (see table \ref{tableTest}). Thus, the
model found by model merging can be regarded generally better than the
bigram model.

\begin{table}
\caption{Number of states and Log Perplexity for the derived models and an
	additional, previously test
	part, consisting of 9,784 words. (a) standard bigram model, (b)
	constrained model merging (first experiment), (c) model merging
	starting with a
	bigram model(second experiment)}
\label{tableTest}
\hrule
\begin{center}
\begin{tabular}{l|ccc}
	& (a)		& (b)		& (c) \\
\hline
	& 		& model		& MM start \\
type	& bigrams	& merging	& with bigrams \\
\hline
\# states & 1,440	& 113		& 113 \\
Log PP  & 2.78	& 2.41		& 2.39 \\
\end{tabular}
\end{center}
\hrule
\end{table}

\subsection{Improvements}

The derivation of the optimal model took about a week although the size
of the training part was relatively small. Standard speech applications
do not use 14,000 words for training as we do in this experiment, but
100,000, 200,000 or more. It is not possible to start with a model of
100,000 states and to successively merge them, at least it is not
possible on today's machines. Each step would require the test of
$\approx 10^9$ merges. 

\begin{figure*}
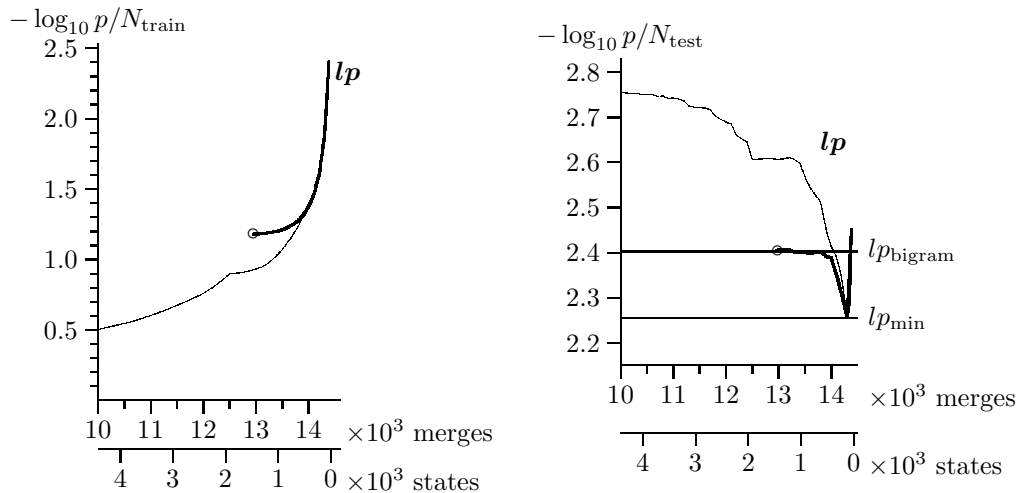

\hrule
\bigskip
\hspace*{0pt}
\raise 10mm\hbox{\beginpicture
\setcoordinatesystem units <.7mm,1.3mm>
\setplotarea x from 100 to 146, y from 0 to 36.5
\axis bottom ticks 
	withvalues 10 11 12 13 14 /
	from 100 to 140 by 10 /
\axis bottom ticks short from 105 to 135 by 10 /
\put {$\times 10^3$ merges} <0mm,-2.5mm> [lt] at 147 0
\axis bottom shiftedto y=-5 ticks
	withvalues 4 3 2 1 0 /
	at 104.23 114.23 124.23 134.23 144.23 / /
\put {$\times 10^3$ states} [lt] <0mm,-2.5mm> at 147 -5
\axis left ticks
	withvalues 0.5 1.0 1.5 2.0 2.5 /
	at 7.21 14.42 21.63 28.84 36.05 / /
\axis left ticks short
	at 1.44  2.88  4.33  5.77  8.65 10.09 11.54 12.98 15.86 17.31
	   18.75 20.19 23.07 24.52 25.96 27.40 30.28 31.73 33.17 34.61 / /
\put {$-\log_{10} p / N_{\mbox{\scriptsize train}}$} [b] at 100 37.5
\linethickness=1.5mm
\setlinear
\setplotsymbol({\rule{.1pt}{.1pt}})
\plot
	100	 7.2852
	101	 7.4011
	102	 7.5127
	103	 7.6289
	104	 7.7477
	105	 7.8918
	106	 8.0256
	107	 8.1826
	108	 8.3440
	109	 8.5169
	110	 8.6964
	111	 8.8957
	112	 9.0807
	113	 9.2833
	114	 9.5090
	115	 9.7344
	116	 9.9711
	117	10.2040
	118	10.4439
	119	10.6891
	120	10.9831
	121	11.2966
	122	11.6608
	123	12.0597
	124	12.4951
	125	12.9675
	126	13.0218
	127	13.0851
	128	13.1711
	129	13.2827
	130	13.4380
	131	13.6557
	132	13.9411
	133	14.3525
	134	14.8646
	135	15.4545
	136	16.1519
	137	16.9282
	138	17.7940
	139	18.7785
	140	19.9588
	141	21.3818
	142	23.4625
	143	27.2294
	143.87	34.7349
/
\setplotsymbol({\rule{.1pt}{.1pt}})
\put {$\circ$} at 129.47 17.0087
\setplotsymbol({\rule{.8pt}{.8pt}})
\plot
	129.47	17.0087
	130	17.0177
	131	17.0757
	132	17.1402
	133	17.2127
	134	17.3278
	135	17.4762
	136	17.6861
	137	17.9817
	138	18.4022
	139	18.9881
	140	19.8239
	141	21.0967
	142	23.1166
	143	26.9635
	143.86	34.5472
/
\put {\boldmath$lp$} [lt] at 145 34.7349
\endpicture}
\hspace*{10pt}
\mbox{\beginpicture
\setcoordinatesystem units <.7mm,60mm>
\setplotarea x from 100 to 145, y from 2.15 to 2.83
\axis bottom ticks 
	withvalues 10 11 12 13 14 /
	from 100 to 140 by 10 /
\axis bottom ticks short from 105 to 135 by 10 /
\put {$\times 10^3$ merges} [lt] <0mm,-2.5mm> at 147 2.15
\axis bottom shiftedto y=2 ticks
	withvalues 4 3 2 1 0 /
	at 104.23 114.23 124.23 134.23 144.23 / /
\put {$\times 10^3$ states} [lt] <0mm,-2.5mm> at 147 2
\axis left ticks 
	numbered from 2.2 to 2.8 by .1 /
\put {$-\log_{10} p/N_{\mbox{\scriptsize test}}$} [b] at 100 2.85
\linethickness=1.5mm
\setlinear
\setplotsymbol({\rule{.1pt}{.1pt}})
\plot
	100.00	 2.7558
	101.00	 2.7542
	102.00	 2.7527
	103.00	 2.7534
	104.00	 2.7506
	105.00	 2.7514
	106.00	 2.7503
	107.00	 2.7469
	108.00	 2.7475
	109.00	 2.7429
	110.00	 2.7431
	111.00	 2.7414
	112.00	 2.7365
	113.00	 2.7255
	114.00	 2.7219
	115.00	 2.7218
	116.00	 2.7206
	117.00	 2.7171
	118.00	 2.7030
	119.00	 2.6964
	120.00	 2.6898
	121.00	 2.6853
	122.00	 2.6598
	123.00	 2.6532
	124.00	 2.6451
	125.00	 2.6072
	126.00	 2.6074
	127.00	 2.6082
	128.00	 2.6087
	129.00	 2.6082
	130.00	 2.6070
	131.00	 2.6086
	132.00	 2.6107
	133.00	 2.6069
	134.00	 2.5994
	135.00	 2.5695
	136.00	 2.5438
	137.00	 2.5283
	138.00	 2.5119
	139.00	 2.4530
	140.00	 2.4129
	141.00	 2.3908
	142.00	 2.3431
	143.00	 2.2587
	143.10	 2.2548
	143.20	 2.2595
	143.30	 2.2620
	143.40	 2.2684
	143.50	 2.2712
	143.60	 2.2979
	143.70	 2.3196
	143.80	 2.3517
	143.87	 2.4516 
/
\put {\boldmath$lp$} [lb] <0pt,1mm> at 138 2.6
\setplotsymbol({\rule{.8pt}{.8pt}})
\plot
	129.47	 2.4030
	130.00	 2.4065
	131.00	 2.4065
	132.00	 2.4067
	133.00	 2.4014
	134.00	 2.4019
	135.00	 2.4008
	136.00	 2.3984
	137.00	 2.4003
	138.00	 2.4014
	139.00	 2.3924
	140.00	 2.3882
	141.00	 2.3514
	142.00	 2.3051
	143.00	 2.2592
	143.86	 2.4516
/
\linethickness=.4pt
\putrule from 100 2.4030 to 145 2.4030
\put {$lp_{\mbox{\scriptsize bigram}}$} [l] at 147 2.4030
\put {$\circ$} at 129.83 2.4030
\putrule from 100 2.2548 to 145 2.2548
\put {$lp_{\mbox{\scriptsize min}}$} [l] at 147 2.2548
\endpicture}
\bigskip
\hrule
\caption{Log Perplexity of training and test parts when starting with a
bigram model. The starting point is indicated with $\circ$, the curves
of the previous experiment are shown in thin lines.}
\label{figBigramMerge}
\end{figure*}

In the previous experiment, we abandoned the same-output constraint
after 12,500 merges to keep the influence on the final result as small
as possible. It can not be skipped from the beginning because somehow
the time complexity has to be reduced. But it can be further retained,
until no further merge under this constraint is possible. This yields a
bigram model. The second experiment uses the bigram model with 1,440
states as its starting point and imposes no constraints on the merges.
The results are shown in figure \ref{figBigramMerge}. 

We see that the perplexity curves approach very fast their counterparts
from the previous experiment. The states differ from those of the
previously found model, but there is no difference in the number of
states and corpus perplexity in the optimal point. So, one could in
fact, at least in the shown case, start with the bigram model without
loosing anything. Finally, we calculate the perplexity for the
additional test part. It is 2.39, thus again lower than the perplexity
of the bigram model (see table \ref{tableTest}). It is even slightly
lower than in the previous experiment, but most probably due to random
variation.

The derived models are not in any case equivalent (with respect to
perplexity), regardless whether we start with the trivial model or the
bigram model. We ascribe the equivalence in the experiment to the
particular size of the training corpus. For a larger training corpus,
the optimal model should be closer in size to the bigram model, or even
larger than a bigram model. In such a case  starting with bigrams does
not lead to an optimal model, and a trigram model must be used.

\section{Conclusion}
\label{secConclusion}

We investigated model merging, a technique to induce Markov models from
corpora. The original procedure is improved by introducing constraints
and a different initial model. The procedures are shown to be applicable
to a transliterated speech corpus. The derived models assign lower
perplexities to test data than the standard bigram model derived from
the same training corpus. Additionally, the merged model was much
smaller than the bigram model. 

The experiments revealed a feature of model merging that allows for
improvement of the method's time complexity. There is a large initial
part of merges that do not change the model's perplexity w.r.t.\ the
test part, and that do not influence the final optimal model. The time
needed to derive a model is drastically reduced by abbreviating these
initial merges. Instead of starting with the trivial model, one can
start with a smaller, easy-to-produce model, but one has to ensure that
its size is still larger than the optimal model.

\section{Acknowledgements}

I would like to thank Christer Samuelsson for very useful comments on
this paper. This work was supported by the Graduiertenkolleg
Kognitionswissenschaft, Saarbr\"ucken.

\end{document}